\documentclass[a4paper,fleqn]{cas-sc}
 \usepackage{tikz}
 \usepackage{amsmath}
 \usetikzlibrary{shapes.geometric, arrows.meta, positioning, backgrounds, fit, calc}
\usepackage[authoryear]{natbib}
\usepackage{subcaption}
\usepackage{algorithm}
\usepackage{algpseudocode}
\def\tsc#1{\csdef{#1}{\textsc{\lowercase{#1}}\xspace}}
\tsc{WGM}
\tsc{QE}
\tsc{EP}
\tsc{PMS}
\tsc{BEC}
\tsc{DE}

\begin{document}
\let\WriteBookmarks\relax
\def\floatpagepagefraction{1}
\def\textpagefraction{.001}

\shorttitle{GREAT-ESDD}

\shortauthors{Lischka et~al.}

\title [mode = title]{GREAT-ESDD: Graph Edge Attention Network for Effectively Steered Delivery Decisions}

\author[1]{Attila Lischka}[
                        orcid=0009-0003-1371-5996]

\cormark[1]

\ead{lischka@chalmers.se}

\affiliation[1]{organization={Electrical Engineering, Chalmers University of Technology},
    city={Gothenburg},
    country={Sweden}}

\affiliation[2]{organization={Civil and Environmental Engineering, University of Washington},
    city={Seattle},
    country={USA}}

\author[1]{Ze Zhou}[orcid=0000-0003-1929-9850]

\author[2]{Xuegang (Jeff) Ban}

\author[1]{Balázs Kulcsár}[orcid=0000-0002-3688-1108]

\cortext[cor1]{Corresponding author}

\begin{abstract}
Online, same-day delivery operations require operators to constantly revise their delivery routes. 
As the routing cost increases with higher customer numbers, out-of-home (OOH) delivery options can be an effective method to allow companies to increase or maintain customer service numbers and satisfaction while decreasing delivery cost.
In this work, we develop a dynamic pricing-and-routing framework for such an online ordering scenario.

To achieve this, we first develop a Graph-Edge-Attention-Network-based (GREAT-based) neural combinatorial optimization framework to solve the capacitated vehicle routing problem (CVRP) of varying instance sizes and starting conditions. 
This allows us to create an effective model to reoptimize partially executed routes under dynamically changing conditions during operations. 
We demonstrate that our GREAT-based solver is fast and achieves competitive solution quality compared to traditional baselines.
Building on the capabilities of our neural solver which can quickly solve several CVRP instances in parallel in batches, we then create a pricing-and-routing framework.
We dynamically solve several candidate routing problems that arise based on an incoming customer's decision for either home or one of the OOH delivery options. 
This serves two purposes: 
First, it allows us to obtain an estimate of the delivery cost increase based on the customer choice. 
Since our model is fast, it allows us to utilize this estimate to dynamically provide the customer with prices for the different delivery options.
Second, we can directly obtain the new route the delivery vehicle should follow once the customer makes a final decision. 
By selecting the dynamic prices in a way that trades off delivery fees and routing costs, we are able to steer customers towards OOH delivery if home delivery would result in a high delivery cost increase. 
For effective steering, an estimate of the customer preferences for delivery options is required.
We show that we can learn those preferences online with Thompson sampling, a Bayesian bandit approach, using a Laplace-approximated posterior to recover the missing parameters in order to adjust our pricing strategy.

We call the resulting framework GREAT-ESDD (GREAT for Effectively Steered Delivery Decisions).
Compared to a traditional routing solver, we can increase profit by around 11\% under a similar decision time budget.
Further, we show that we can successfully estimate the customer preferences online, allowing us to achieve a higher profit than a static pricing strategy after only 5 days of simulated operations.

\end{abstract}

\begin{keywords}
Neural Combinatorial Optimization \sep Dynamic Pricing \sep Out-of-home Delivery \sep Reinforcement Learning \sep Bandit Learning \sep Thompson Sampling
\end{keywords}

\maketitle

\section{Introduction}
Last-mile parcel delivery logistics are comparatively expensive \citep{boysen2021last, PAHWA2022102783}.
Each additional stop lengthens a delivery tour, and delivery visits that lie far from the current route are disproportionately cost-intensive.
An effective way to reduce delivery costs is out-of-home (OOH) delivery, where, in contrast to attended home delivery, parcels are brought to a number of fixed delivery locations such as parcel shops or locker boxes.
OOH deliveries ideally permit the operator to deliver several parcels for multiple customers at once in a single stop \citep{MANCINI2021105361}.

However, instead of simply depriving the customer from the home-delivery option and forcing them to use shared OOH delivery, \emph{pricing} can be a more elegant way to steer customer choices towards OOH delivery.
By offering a variety of delivery options at different price points, the operator can nudge customers to select delivery options that are cheap to serve \citep{campbell2006incentive, yang2016choice}.
Making an informed, beneficial pricing decision that steers the customer towards cheap-to-serve delivery options requires that we can estimate how expensive a certain delivery option is for the operating company. 
This cost is the marginal increase in routing cost from inserting a possible delivery stop based on the vehicle's current position and its already planned tour. 
Computing this cost corresponds to solving a vehicle routing problem (VRP). 
In an online setting, where customers can order continuously during ongoing operation, this means solving a VRP for every delivery option for each arriving customer while the customer waits at a checkout page for the delivery options at different price points. 
Consequently, we need to solve the VRPs quickly, resulting in a setting where classical exact or metaheuristic routing solvers are too slow. 
Moreover, once the customer has made a decision, we can update the delivery vehicle's route in real time using the computed solution. 

While many works exist that use machine learning to solve VRPs (e.g., \citealp{kool2018attention, joshi2019efficientgraphconvolutionalnetwork, nazari2018reinforcement}), most of those papers focus on settings with a fixed number of nodes and ``empty'' starting conditions (i.e., the vehicle starts in the depot fully loaded).
A route is computed once, and the vehicle executes it without any new customers entering or stochastic events happening in the system.
A dynamic setting, on the other hand, where new, additional customers come into the system over time is comparatively underexplored.
This is a limitation in settings such as the previously described pricing context where we need a solver that can support pricing decisions for partially executed routes and a vehicle that is already en route and (partially) loaded.

In this work, we propose a deep-reinforcement-learning-based framework to solve a mixed-size version of the capacitated vehicle routing problem (CVRP) with arbitrary starting state. 
We use this neural solver, which is based on the GRaph Edge ATtention network (GREAT), inside an online pricing-and-routing framework.
The trained solver serves two roles at inference: it acts as a routing cost estimator during pricing to steer customer decisions and, moreover, it produces the route the delivery vehicle follows after each state change.
GPU-based batching enables us to solve the VRPs for all possible delivery options efficiently in parallel in real time. 
The resulting framework is called Graph Edge Attention Network for Effectively Steered Delivery Decisions (GREAT-ESDD).
The resulting operational loop can be seen in Figure~\ref{fig:pipeline}.

In particular, our contributions are:
\begin{itemize}
    \item \textbf{A fast neural solver for partially executed routes.} We introduce a GREAT-based CVRP model that can solve CVRP instances of mixed sizes. 
    The proposed model further accommodates routes that start at any node, not only the depot, with a partially loaded vehicle, thereby enabling the representation of partial solutions that must be extended and finalized. Through its robustness to different sizes and starting points, the trained model can be used to solve dynamic CVRPs during inference by encoding the partial solution and new customers as a new graph and iterative reoptimization. We show that the resulting trained model is fast and accurate enough to be called inside an online decision loop. It is competitive with a state-of-the-art heuristic solver under matched decision-time budgets. Further, it generalizes to instance sizes beyond those encountered during training, which allows us to solve problems with fluctuating numbers of customers over an operating day.
    \item \textbf{A pricing-and-routing framework.} We build a pricing-and-routing framework named GREAT-ESDD where customers iteratively enter the system and are offered prices for different home and out-of-home delivery options. For every arriving customer, our trained CVRP model solves the candidate routing problems of all delivery options in parallel, which yields the routing cost increase of each option. We use these increases to dynamically compute how to select the offered prices. After the customer makes a choice, we can directly obtain the new route the delivery vehicle should follow from the already solved candidate routing problem. Under similar decision-time budgets, GREAT-ESDD increases profit by around 11\% compared to a traditional routing solver.
    \item \textbf{Online learning of customer preferences.} We apply a multinomial logit (MNL) model to model the customer utilities and learn their parameters online with Thompson sampling in a bandit setting in order to maximize profit. We show that the parameters are approximately recovered and that the learned dynamic pricing outperforms the best flat pricing strategy within 5 days of simulated operations.
\end{itemize}

The remainder of this work is organized as follows:
Section~\ref{sec:related_work} introduces the related literature in learning-based routing and pricing-and-routing settings. 
Section~\ref{sec:problem_setting} describes the dynamic pricing-and-routing problem, including the customer choice model and the operator's objective. Section~\ref{sec:methodology} gives an overview of GREAT-ESDD and introduces our neural solver, the pricing-optimization framework and the online preference learning setting.
Section~\ref{sec:experiments} describes the experimental setup and Section~\ref{sec:results} reports the results before we discuss limitations in Section~\ref{sec:limitations} and conclude in Section~\ref{sec:conclusion}.

\section{Related work}

\label{sec:related_work}

\subsection{Machine learning for static routing problems}
A pioneering work using supervised machine learning to solve the Traveling Salesman Problem using pointer networks was \cite{vinyals2015pointer}. Subsequently, papers started using reinforcement learning to solve TSP to overcome the need for labeled data \citep{bello2017neural, khalil2017learning}.
Early works that started to solve the CVRP with such learning based methods are \cite{nazari2018reinforcement, kool2018attention}. 
Their work laid the foundation for the rapid growth of machine-learning-based VRP heuristics.
Besides improving performance \citep{kwon2020pomo}, recent developments focus on solving larger and larger instances \citep{drakulic2023bqnco, ye2024glop}, creating routing foundation models that can solve several VRP variants \citep{zhou2024mvmoe, drakulic2025goal} and routing on more complicated and realistic distributions \citep{kwon2021matrix, lischka2025greatarchitectureedgebasedgraph, son2026rrnco, yi2026radar}.

A detailed overview of developments for learning for routing can be found in \cite{ZHOU2025104278}.

Our solver builds directly on this line of work, however, it has two adaptations to fit our setting: first, it is trained on a range of instance sizes (instead of a single fixed instance size like it is common for other learning-based routing solvers) to allow us to tackle different numbers of active customers over the course of a day. Second, it is trained to start from arbitrary nodes with arbitrary vehicle loads to capture the essence of partially solved instances that need to be re-solved. This is in contrast to existing work that typically assumes an empty vehicle starting at the depot.

\subsection{Machine learning for dynamic routing problems} \label{sec:dynamic_routing_literature}
While the vast majority of machine-learning-based work focuses on solving static VRPs, there are some papers tackling settings with dynamic components.
An early and pioneering work is \cite{yu_online_vr}. Using an actor-critic reinforcement learning framework, the authors train a structure2vec model \citep{dai2016discriminative} jointly with a pointer network to generate solutions for an electric-vehicle-based CVRP variant with dynamic customer arrival. In their setting, several vehicles are routed simultaneously and independently, by creating a separate graph representation of the current VRP state for each vehicle. This distributed approach is fast but does not optimize the overall setting holistically. Another key distinction compared to static papers like e.g. \cite{kool2018attention, kwon2020pomo} is the handling of constraints. While those papers enforce side constraints (such as vehicle capacity) by masking operations during solution generation (preventing the vehicle from making invalid moves), \cite{yu_online_vr} use soft constraints whose violation is penalized in the objective function.
\cite{joe2020deep} also tackle a dynamic VRP setting. They formulate the problem as an MDP and train a simple neural network in the form of a multilayer perceptron for value function approximation. Those values are used in a simulated annealing framework to evaluate different re-routing options whenever a new customer enters the system. Consequently, machine learning is used as guidance in this setting only and not in an end-to-end fashion that directly constructs routes.

\cite{NEURIPS2021_c6a01432} introduce a two-stage hierarchical framework to tackle their dynamic setting consisting of 144 timeslots. At each timeslot, the upper-level agent decides if potential new customers that have arrived during the previous timeslot should be released into the system and a new route should be constructed. Alternatively, the new customers are cached and only released during a later timeslot. Once the new customers are released, a new route is constructed using a simple greedy heuristic and, subsequently, the route is improved with learned improvement operators (such as swapping nodes in the solution tour).

\cite{baty2024combinatorial} interpret their dynamic problem at each timestep as a prize collecting TSP with time windows. They train a neural network to predict the prizes (corresponding to future demand) of the TSP variant and consequently solve the problem with a solver. 

\cite{KADYROV2025100351} consider a dynamic setting with stochastic node demands and travel costs. The customers themselves, however, are known from the beginning and arranged in a synthetic grid-shaped environment which is processed by a GNN-based model trained with RL.

Finally, \cite{sultana2021fastapproximatesolutionsusing} also considers dynamic customer arrivals and uses Q-learning to train a multilayer perceptron to predict scores for feasible (i.e. non-VRP-constraint violating) vehicle-customer pairs.
A vehicle is assigned one customer at a time only, with a new assignment round being triggered each time a vehicle becomes free. A soft assignment mechanism is used to ensure conflicts (such as assigning several vehicles to the same customer) are avoided.

In contrast to these works, we do not try to learn the dynamic aspect of our problem. Training a dynamic model would require us to model a time component and perfectly match the customer arrival rates during training and inference. Since the exact customer count in different instances can quickly diverge over the day, batching several instances is not possible, complicating processing during both training and inference. Instead, we train a size and start-condition robust model and achieve dynamic behavior through iterative re-optimization during inference. This sacrifices the ability to anticipate future arrivals which we will discuss in Section~\ref{sec:limitations} but makes the solver cheap to train and transferable between distributions during inference.

\subsection{Pricing and routing}

Combining pricing and routing decisions is an active area of research, especially in the context of attended home delivery and time slot booking \citep{WAMUTH2023801}.
An early work that investigates how to steer customers to make decisions that are inexpensive to serve is \cite{campbell2006incentive}.
In \cite{yang2016choice}, a multinomial logit customer choice model is estimated from historical booking data to determine the required incentive to influence a customer decision. 
At the tactical level, \cite{klein2019differentiated} integrate differentiated time slot pricing with customer choice and routing considerations. \cite{strauss2021dynamic} extend dynamic pricing to regular and flexible delivery time slots, accounting for the routing benefits of customers accepting greater delivery-time flexibility.

To support pricing decisions, opportunity cost approximations have been used to account for both incremental delivery costs and the impact of current bookings on future demand acceptance \citep{klein2018model, KOCH2020633}. 
\cite{yang2017approximate} and \cite{vinsensius2020dynamic} develop approximate dynamic programming approaches that account for future orders when determining delivery slot prices or incentives. 
\cite{fleckenstein2025concept} provide a general foundation for opportunity costs in integrated demand management and vehicle routing, clarifying the distinction between displacement costs and marginal service costs. 
Using an explicit routing approach, \cite{abdollahi2023demand} maintain routes containing committed orders and forecast orders without assigned time windows, and reoptimize these routes during the booking horizon to guide dynamic pricing.
\cite{ulmer2020dynamic} uses a value function approximation in a pricing-and-routing framework for same day delivery to price different delivery deadlines and \cite{KLEIN2023860} use dynamic demand management and online tour planning.

More similar to our research direction, \cite{MANCINI2021105361} combine private (home) and shared (OOH) delivery options to trade off quality of service for the customers against the company's operational cost. \cite{galiullina2024demand} formulate a two-stage stochastic program that selects customers to incentivize toward pickup point delivery and determines routes after their uncertain acceptance decisions are revealed. 
\cite{janinhoff2024multitrip} develop a multitrip routing framework with delivery location options and evaluate how OOH product design and home-delivery surcharges affect operational costs and profitability.
\cite{akkerman2025learning} use a convolutional neural network to predict routing costs and use those in a pricing framework for home and OOH deliveries.
\cite{ZHOU2026104953} propose a framework that considers both time slots and OOH delivery options and develop a hybrid adaptive large neighborhood search to solve the resulting static mixed-integer linear program. 
In a dynamic extension with sequential customer arrivals, \cite{zhou2026assortment} jointly optimize delivery-option assortments and dynamic prices for home and OOH delivery, using approximate dynamic programming to estimate opportunity costs.

Our work differs from this literature in how we compute the marginal routing cost increase and perform the online routing decision. Opportunity cost models \citep{klein2018model, KOCH2020633} and value functions \citep{ulmer2020dynamic} approximate the cost of accepting an order including its impact on future demand, i.e., they account for displacement costs in addition to marginal service costs \citep{fleckenstein2025concept}, while learned cost predictors \citep{akkerman2025learning} approximate routing costs from the current state. Instead, we compute the actual routing cost increase by solving the candidate routing problems. As discussed in Section~\ref{sec:dynamic_routing_literature}, our solver sacrifices the ability to anticipate future customer arrivals, so the resulting estimate is myopic. We only partially compensate for this with a rectifier (Section~\ref{sec:rectifying_factor}). In return, the marginal service cost is obtained from a heuristic solution of the current routing problem rather than from an approximation. Solving several problems, one for each delivery option, at decision time is only practical within sub-second budgets with a fast, batched solver. Computing actual VRP solutions also allows us to directly update our vehicle's route after a customer has made a decision. Notably, this would not be possible when using an opportunity cost, value function or learned cost predictor alone.

\section{Problem description}
\label{sec:problem_setting}

\subsection{Operational setting}

We consider a delivery company that serves customers from a single central depot with a single vehicle of capacity $Q$.
Each customer $i$ places an order with an integer demand $q_i \in \{1, \dots, \bar q\}$.
If the cumulative demand of the customers on the planned route exceeds the vehicle's remaining load, the vehicle has to return to the depot to reload before it can serve further customers, so that the routing problem at any point in time is a CVRP.
Traveling from location $u$ to location $v$ incurs a cost $c_{uv}$, for example the travel time, which is not necessarily symmetric.
All goods are assumed to be interchangeable.
Optionally, customer demands can be stochastic in the sense that the realized demand of a customer, which is revealed when the vehicle arrives, may be lower than the ordered demand.
The freed load can then be used to serve further customers without returning to the depot.
This situation arises, for instance, in pick-up operations in which goods are collected rather than delivered, such as customers returning parcels to an online shop.
For simplicity, we nevertheless refer to all services as deliveries throughout this work.

Customers arrive sequentially over the course of a day.
Some of them order before operations begin, for example overnight, while the remaining customers order dynamically while the vehicle is already en route.
Neither the number of customers nor their arrival times, locations and demands are known to the operator in advance.
Whenever a customer orders, the operator offers a menu of delivery options together with a price for each option.
The customer can choose home delivery, choose out-of-home delivery to one of $K$ fixed OOH locations such as parcel shops or locker boxes, or opt-out and not place the order.
Prices are dynamic and can differ between customers, and they must be determined in real time, since the customer waits for the price menu during checkout.
The customer's choice depends on the offered prices and, for OOH delivery, on the distance between the customer's home and the respective OOH location, as formalized in Section~\ref{sec:customer_choice_model}.

The goal of the operator is to maximize the expected profit over the operating day, defined as the sum of all prices paid by customers minus the cost of the route the vehicle travels over the day.
The price menu is the operator's lever to achieve this goal: by steering customers towards delivery options that are cheap to serve, the operator trades off collected fees against routing cost.

\subsection{Customer choice model}
\label{sec:customer_choice_model}
The operator offers a price for each delivery option.
In the following, we describe how customers choose either one of the delivery options or opting-out of the delivery process altogether.

We assume that prices are taken from the discrete grid
\begin{equation}
    \mathcal{G} = \{p_{\min},\, p_{\min} + \delta,\, p_{\min} + 2\delta,\, \dots,\, p_{\max}\},
\end{equation}
where $p_{\min}$ and $p_{\max}$ denote the lowest and highest price and $\delta$ the price increment.

Customers convert these prices into \emph{utilities}, using the following utility model:

\begin{align} \label{eq:utility1}
    U_h(p_h) &= \alpha_h - \beta_h\, p_h, \\
    U_{{OOH}_k}(p_{{OOH}_k}, d_{{OOH}_k}) &= \alpha_{OOH} - \beta_{OOH}\, p_{{OOH}_k} - \gamma\, d_{{OOH}_k},
        \qquad k = 1,\dots,K, \\
    U_\emptyset &= 0. \label{eq:utility2}
\end{align}
where $U_h$ is the utility of home delivery, $U_{OOH_k}$ the utility of delivery to OOH location $k$ and $U_\emptyset$ the utility of opting-out.
Note that the utility for home delivery is computed using only the price $p_h$ as an input, whereas OOH delivery has, additionally to the price input $p_{OOH_k}$, another input $d_{OOH_k}$ that corresponds to the distance, for example measured as travel time, between the OOH location $k$ and the customer.
Moreover, the opt-out utility is constant and corresponds to 0.
The utilities are defined by five parameters: $(\alpha_h, \beta_h, \alpha_{OOH}, \beta_{OOH}, \gamma)$. Here, $\alpha_h$ is the base utility for home delivery and $\beta_h$ the price sensitivity for home delivery.
Further, we have $\alpha_{OOH}$ as the base utility for OOH delivery, $\beta_{OOH}$ the price sensitivity for OOH delivery and $\gamma$ the sensitivity for the distance of an OOH location to a customer's home.

Given those customer utilities, we use an MNL  model to obtain the probability of a customer choosing an option $j \in \mathcal J = \{h, OOH_1, \dots, OOH_K, \emptyset\}$ from the different options available to the customer. For an offered combination of delivery prices $\mathbf p = (p_h, p_{{OOH}_1},\dots,p_{{OOH}_K})$ and a distance vector $\mathbf d = (d_{{OOH}_1},\dots,d_{{OOH}_K})$ that specifies the distances from the customer's home to each OOH location, the choice probability of option $j \in \mathcal J$ is

\begin{equation}
    \mathbb P(j \mid \mathbf p, \mathbf d)
    = \frac{\exp(U_j)}{\sum_{j' \in \mathcal J} \exp(U_{j'})}. \label{eq:cust_prob}
\end{equation}
Note that this is a slight abuse of notation and $U_j$ refers to the utility model introduced above in \eqref{eq:utility1} -- \eqref{eq:utility2} that is a function having different inputs depending on the option.

\subsection{The dynamic pricing-and-routing problem}
\label{sec:operator_price_decision}
We index customers by $i = 1, 2, \dots$ in the order of their arrival at times $t_1 \le t_2 \le \dots$, where the total number of customers per day is not known in advance.
Customer $i$ is characterized by the location of their home, their demand $q_i$ and the vector $\mathbf d_i = (d_{i,OOH_1}, \dots, d_{i,OOH_K})$ of distances between their home and the OOH locations.
When customer $i$ arrives, the system is in the routing state $\tau_i$, which comprises the accepted but still unserved customers together with their delivery locations and demands, as well as the vehicle's current position and remaining load.

At each arrival, the operator makes two decisions.
First, it offers a price menu $\mathbf p_i = (p_{i,h}, p_{i,OOH_1}, \dots, p_{i,OOH_K}) \in \mathcal G^{K+1}$, and the customer chooses an option $j_i \in \mathcal J$ according to \eqref{eq:cust_prob} with $\mathbf p = \mathbf p_i$ and $\mathbf d = \mathbf d_i$.
Second, if $j_i \neq \emptyset$, the customer is added to the set of accepted customers with the delivery location given by $j_i$, and the operator determines an updated route that serves all accepted and unserved customers, starting from the vehicle's current position and load.
The vehicle follows this route until the next customer arrives or a realized demand deviates from the ordered demand, at which point the route can be revised.
Once no further customers arrive, the vehicle completes its route and returns to the depot.

Let $C$ denote the total cost of the route traveled by the vehicle over the day and let $p_{i,\emptyset} = 0$.
The operator seeks a pricing-and-routing policy that maximizes the expected daily profit
\begin{equation} \label{eq:daily_profit}
    \mathbb E\Bigl[\,\sum_{i} p_{i, j_i} - \theta\, C\Bigr],
\end{equation}
where $\theta$ is a scaling parameter that maps routing cost to monetary cost, e.g., delivery operation time to dollars, and the expectation is taken over customer arrivals, customer choices and realized demands.

Three features make this problem challenging.
First, the profit contribution of a customer depends on how the chosen delivery option changes the route, which in turn depends on all future customers.
An exact solution therefore requires anticipating future demand and is intractable for realistic instances.
Second, prices must be determined in real time, since the customer waits for the price menu during checkout.
Third, the choice probabilities depend on the preference parameters $(\alpha_h, \beta_h, \alpha_{OOH}, \beta_{OOH}, \gamma)$, which are typically unknown to the operator and have to be learned from observed customer choices while operations are already running.
Section~\ref{sec:methodology} presents our solution approach, which addresses these three challenges.

\section{Solution approach}
\label{sec:methodology}
\begin{figure}
\centering
\scalebox{0.75}{%
\begin{tikzpicture}[
    font=\small,
    node distance=8mm and 12mm,
    >={Stealth[length=2.5mm]},
    startstop/.style={
        rectangle, rounded corners=3mm, draw=black!70, thick,
        fill=gray!12, minimum height=9mm, inner sep=2.5mm, align=center, text width=45mm},
    process/.style={
        rectangle, rounded corners=1mm, draw=blue!50!black, thick,
        fill=blue!8, minimum height=9mm, inner sep=2.5mm, align=center,
        text width=52mm},
    processend/.style={
        rectangle, rounded corners=1mm, draw=blue!50!black, thick,
        fill=blue!8, minimum height=9mm, inner sep=2.5mm, align=center,
        text width=42mm},
    pricing/.style={
        rectangle, rounded corners=1mm, draw=orange!70!black, thick,
        fill=orange!12, minimum height=9mm, inner sep=2.5mm, align=center,
        text width=90mm},
    decision/.style={
        diamond, aspect=2.2, draw=green!40!black, thick,
        fill=green!8, inner sep=1mm, align=center},
    arrow/.style={->, thick, draw=black!75},
    lab/.style={font=\footnotesize\itshape, fill=white, inner sep=1pt}
]
 
\node (start) [startstop] {\textbf{Night before operations}\\ vehicle idle at the depot};
 
\node (init) [process, below=7mm of start] {Initialize planned route:\\ tour over depot $+$ OOH lockers\\ (active customer set empty)};
 
\node (order) [process, below=7mm of init] {Customer $i$ places an order at time $t_i$};
 
\node (opdec) [decision, below=7mm of order] {Vehicle already\\ en route?};
 
\node (drive) [process, right=18mm of opdec, text width=44mm] {Vehicle follows planned route until $t_i$:\\ serve stops, update position and load,\\ remove served customers};
 
\node (delta) [pricing, below=8mm of opdec] {\textbf{Marginal routing cost:} for every delivery option $j \in \{\text{home},\ \text{OOH}_1,\dots,\text{OOH}_K\}$ compute cost increase $\Delta C_j$ of inserting $j$ into the route, given current vehicle position and load};
 
\node (price) [pricing, below=8mm of delta] {\textbf{Price optimization:} choose fees\\[1mm] $\displaystyle (p_h, p_{OOH_1},\dots,p_{OOH_K}) \in \arg\max_{p}\ \sum_{j} P(j\mid p)\,\bigl(p_j - \theta\,\Delta C_j\bigr)$\\[1mm] with choice probabilities $P(j \mid p)$ from the (learned) customer preference model};
 
\node (offer) [pricing, below=8mm of price] {Offer menu to customer $i$:\\ home delivery at $p_h$, locker OOH$_k$ at $p_k$};
 
\node (choice) [decision, below=8mm of offer] {Customer's\\ choice};
 
\node (accept) [process, below=7mm of choice] {Add customer to active set;\\ update planned route and its cost};
 
\node (more) [decision, below=7mm of accept] {More orders\\ expected?};
 
\node (finish) [processend, right =22mm of more] {Vehicle serves all remaining stops\\ and returns to the depot};
 
\node (end) [startstop, above=8mm of finish] {\textbf{Day ends:} profit $=$ collected fees $-\ \theta \times$ total routing cost};
 
\draw [arrow] (start) -- (init);
\draw [arrow] (init) -- (order);
\draw [arrow] (order) -- (opdec);
 
\draw [arrow] (opdec) -- node[lab, above] {yes} (drive);
\draw [arrow] (opdec) -- node[lab, right, pos=0.35] {no (pre-order phase)} (delta);
\draw [arrow] (drive.south) |- (delta.east);
 
\draw [arrow] (delta) -- (price);
\draw [arrow] (price) -- (offer);
\draw [arrow] (offer) -- (choice);
 
\draw [arrow] (choice) -- node[lab, right, pos=0.4] {home / OOH$_k$} (accept);
\draw [arrow] (choice.east) -- ++(18mm,0) node[lab, above, pos=0.5] {none} |- (more.east);
\draw [arrow] (accept) -- (more);
 
\draw [arrow] (more.west) -- ++(-48mm,0) node[lab, above, pos=0.5] {yes} |- node[lab, above, pos=0.78] {next customer} (order.west);

\draw [arrow] (more.south) -- ++(0,-4mm) coordinate (dip)
    -- node[lab, above, pos=0.5] {no} (dip -| finish.south)
    -- (finish.south);

\draw [arrow] (finish) -- (end);
 
\begin{scope}[on background layer]
    \node (pricingbox) [draw=orange!70!black, dashed, thick, rounded corners=2mm,
        inner sep=3.5mm, fit=(delta)(price)(offer),
        label={[orange!70!black, font=\footnotesize\bfseries, anchor=south west]north west:Pricing decision (per order)}] {};
\end{scope}
 
\end{tikzpicture}}

\caption{The pricing-and-routing loop of the GREAT-ESDD framework over one operating day. The vehicle follows a planned route. Any time a customer orders, the current routing state is used to compute a routing cost estimate of the delivery options and a price menu is chosen to maximize expected profit. The customer's choice determines the updated route.}
\label{fig:pipeline}
\end{figure}
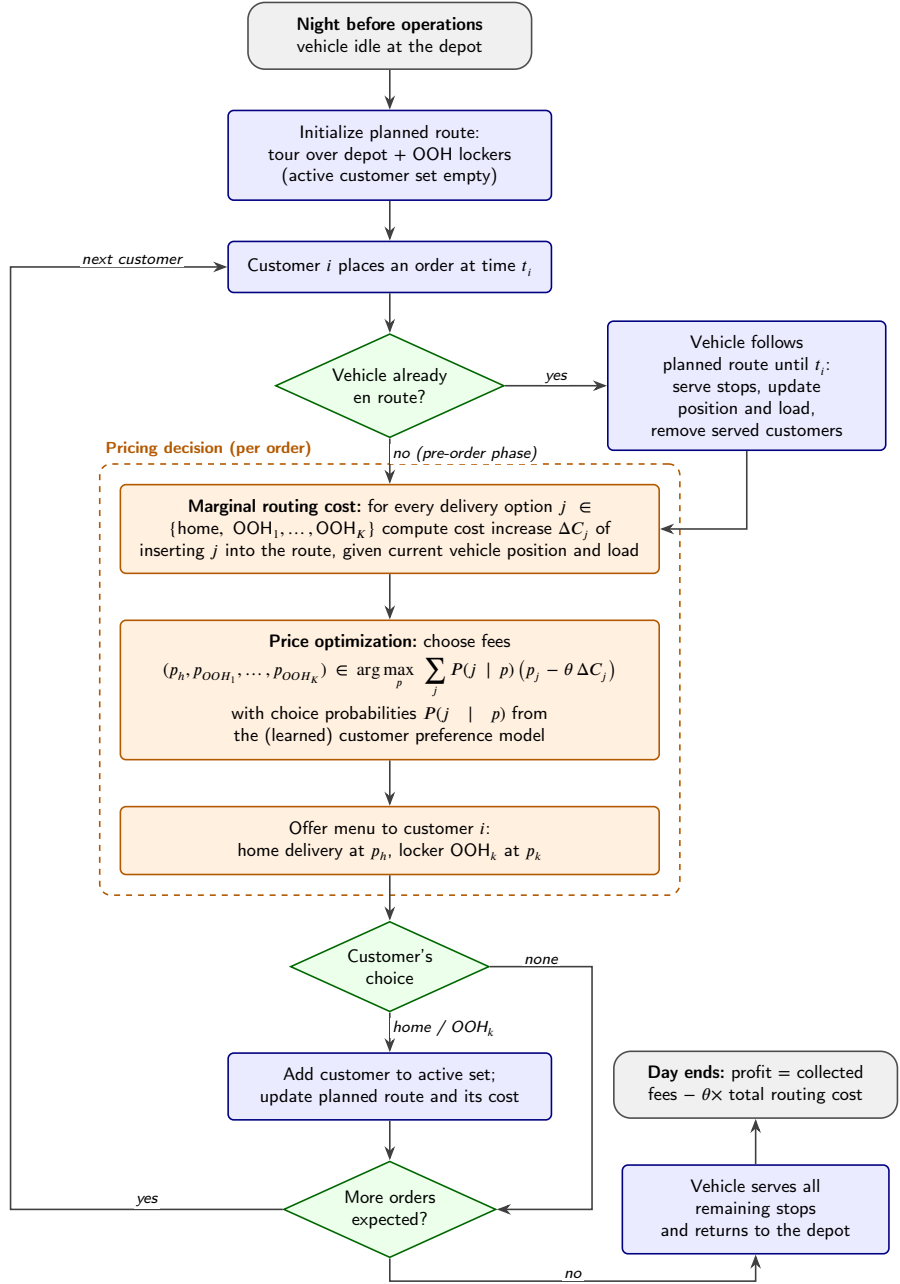

Figure~\ref{fig:pipeline} gives an overview of our solution approach, GREAT-ESDD, over one operating day.
Before operations begin, the vehicle is idle at the depot and its planned route is initialized as a tour that starts at the depot and visits all OOH locations, although none of them has been requested yet.
This reflects the assumption that the OOH locations are visited during the day anyway and prevents the first customers from being charged the full cost of opening the route.
Whenever a customer orders, the pricing decision highlighted in orange in Figure~\ref{fig:pipeline} is executed.
If the vehicle is already en route, it first follows its current route until this point in time, serving the customers on it and updating its position and load.
For the pricing decision, we compute the routing cost increase of every delivery option with the neural CVRP solver introduced in Section~\ref{sec:GREAT} and select the price menu that maximizes the expected profit, as described in Sections~\ref{sec:cost_delta} and~\ref{sec:rectifying_factor}.
The customer preference parameters required for this step are either known or learned online, as described in Section~\ref{sec:online_pref}.
If the customer accepts an option, the vehicle adopts the route that was already computed for this option during pricing, whereas it keeps following its current route if the customer opts-out.
In addition, the route is reoptimized whenever a customer's realized demand is lower than the ordered demand.
Once no further orders arrive, the vehicle completes the remaining route and returns to the depot.
Figure~\ref{fig:overall} illustrates how customer choices and the vehicle route develop over the course of a simulated day.

\begin{figure*}
\centering
\begin{tikzpicture}
\node[anchor=east] (img7) at (0,0) 
    {\includegraphics[page=7, width=3.2cm]{figs/Dynamic_VRP___figures.pdf}};
\node[above] at (img7.north) 
    {\makebox[3.2cm]{\parbox{3.2cm}{\centering Current \\ Solution}}};

\node[anchor=west] (img1) at (0.3, 1.5) 
    {\includegraphics[page=1, width=3.2cm]{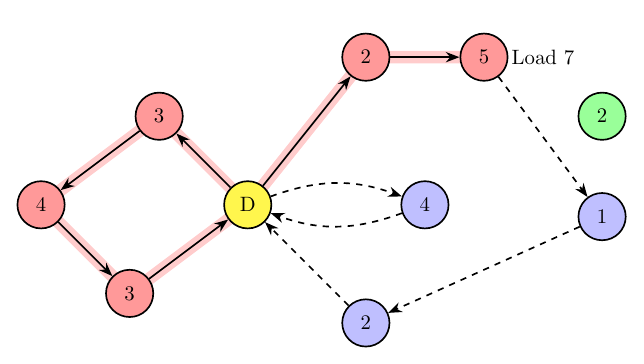}};
\node[above] at (img1.north) 
    {\makebox[3.2cm]{\parbox{3.2cm}{\centering Stochastic \\ Event}}};

\node[anchor=west] (img2) at (4.1, 1.5) 
    {\includegraphics[page=2, width=3.2cm]{figs/Dynamic_VRP___figures.pdf}};
\node[above] at (img2.north) 
    {\makebox[3.2cm]{\parbox{3.2cm}{\centering Graph \\ Representation}}};

\node[anchor=west] (img3) at (7.9, 1.5) 
    {\includegraphics[page=3, width=3.2cm]{figs/Dynamic_VRP___figures.pdf}};
\node[above] at (img3.north) 
    {\makebox[3.2cm]{\parbox{3.2cm}{\centering New \\ Solution}}};

\node[anchor=west] (img4) at (0.3, -1.5) 
    {\includegraphics[page=4, width=3.2cm]{figs/Dynamic_VRP___figures.pdf}};

\node[anchor=west] (img5) at (4.1, -1.5) 
    {\includegraphics[page=5, width=3.2cm]{figs/Dynamic_VRP___figures.pdf}};

\node[anchor=west] (img6) at (7.9, -1.5) 
    {\includegraphics[page=6, width=3.2cm]{figs/Dynamic_VRP___figures.pdf}};

\draw[->, thick] (img7.north east) -- (img1.west);
\draw[->, thick] (img7.south east) -- (img4.west);

\draw[->, thick] (img1.east) -- (img2.west);
\draw[->, thick] (img2.east) -- (img3.west);

\draw[->, thick] (img4.east) -- (img5.west);
\draw[->, thick] (img5.east) -- (img6.west);

\end{tikzpicture}
\caption{Given a current solution, our framework can react to stochastic changes in the environment. Case one (upper branch) occurs when a new customer dynamically enters the system. Case two (lower branch) occurs when a customer has a lower demand than expected. In both cases we construct a new graph representing the current state. It encodes all unserved customers, the depot, the current vehicle location and the current vehicle load. Our neural model processes this graph encoding and outputs a new solution.}
\label{fig:overall_framework}
\end{figure*}

\subsection{A neural solver for CVRP of mixed sizes and arbitrary start states}
\label{sec:GREAT}
The centerpiece of successful operation of our delivery company is efficient and effective routing. 
Since decisions must be made in real time, we require a routing algorithm that is comparatively fast.
Speed is one of the major advantages of neural routing solvers, which is why we aim for such an approach as our routing heuristic.

In our company's CVRP setting, we consider asymmetric distance distributions found in time-based routing problems (arising due to one-way streets and one-way speed limits).
As a consequence, we choose the graph edge attention network (GREAT; \citealp{lischka2025greatarchitectureedgebasedgraph}), which was developed for asymmetric routing problems, as our encoder.
For the decoding, we use the multi-pointer network from \cite{jin2023pointerformer}. 
Similar to those works, we train the model using the reinforcement-learning-based POMO framework \citep{kwon2020pomo} to build a solution tour by adding one node at a time until a complete tour has been constructed. 
We refer the reader to those works for the details of the architecture and briefly describe the construction of a solution as follows.
We model the CVRP as a Markov Decision Process (MDP). The state $s_t$ consists of the instance graph, the partial tour constructed so far, the current node of the vehicle and its remaining load.
An action $a_t$ selects the next node to visit, i.e., an unserved customer or the depot, and the transition appends this node to the partial tour and reduces the remaining load by the node's demand, or resets it to $Q$ at the depot.
The reward of a complete tour is its negative cost, i.e., the negative sum of $c_{uv}$ over all traversed arcs.
Validity of the CVRP, such as respecting vehicle capacity constraints, is ensured by masking actions that lead to invalid moves, i.e., already served customers and customers whose demand exceeds the remaining load.
The policy is a neural network, in our case the aforementioned GREAT encoder plus a multi-pointer network, which produces probabilities representing which node to visit next.
However, we make two fundamental changes compared to existing work that are essential for our setting:

\begin{enumerate}
    \item Unlike existing works that typically focus on training and solving VRPs of a specific size (e.g., CVRP of exactly 100 customers), we train our model on a whole range of sizes (10 to 100 customers). By this, our model is able to accurately solve an emerging CVRP instance during inference, no matter the current number of customers that are present in the system.
    \item Since existing works assume ``empty'' initial solutions, the model is only required to encode and solve instances starting from the depot node with a fully loaded vehicle. To tackle the dynamic nature of our problem, we also allow different start nodes and non-zero vehicle loads during training, allowing the model to learn a more robust strategy for different starting conditions.
\end{enumerate}
The start state enters the model as follows.
In addition to the node demand, we introduce a further node feature that indicates, via a binary encoding, the node where the vehicle is currently located.
The node features are transformed into edge features following \cite{lischka2025greatarchitectureedgebasedgraph} and concatenated with the original edge feature which represents the travel time.
The purely edge-based graph is then passed to the GREAT encoder, so that the start node is identified only through this feature in the encoder.
During decoding, the query of the multi-pointer network is composed of the embedding of the node the vehicle is currently located at, the embedding of the depot, the mean of all node embeddings, an aggregate of the embeddings of already visited nodes, and an embedding of the remaining vehicle load.
The start node itself is marked as visited and therefore enters the query only through this aggregate.
The remaining load of the vehicle additionally initializes the capacity mask, so that only customers whose demand fits into the remaining load can be selected.

We train our model on deterministic instances of different sizes and a variety of starting conditions (vehicle load and starting node). 
By this, we train a state-conditioned neural routing policy. It does not solve the dynamic problem directly. Instead, at each decision epoch, i.e., whenever a customer orders or a realized demand deviates from the ordered demand, it solves the residual static CVRP of all accepted but unserved customers, starting from the vehicle's current position and load. The dynamic behavior of GREAT-ESDD results from this repeated reoptimization.
We show this idea in Figure \ref{fig:overall_framework}. The figure shows a solution to a current routing problem. 
In the stochastic inference setting, it is possible for unexpected events to arise while the vehicle is already en route. 
The two different examples we consider in this work are 1) dynamic, new customers (upper branch in the figure) or 2) unexpected lower customer demand. 
In both cases, we generate the residual CVRP instance that only consists of all currently active customers (already served customers are no longer considered) and the depot. 
We encode this as a complete graph (note that CVRP graphs are complete graphs) where we additionally indicate the current vehicle load, depot node and current vehicle position.
We pass this graph to our robustly trained GREAT model which outputs a new complete solution for the changed setting. 
This new solution can then be followed until the whole CVRP is solved or the next stochastic event occurs.

Our approach of deterministic training decreases the amount of resources required during training considerably. 
Stochastic settings during training would require iterative reconstruction of the changing CVRP graph and subsequent calls to the GREAT encoder. 
This is computationally expensive compared to a single CVRP graph construction and a single GREAT encoder call. 
Furthermore, we perform the graph construction on a CPU while the handling of all neural components is run on GPUs, effectively slowing down the overall pipeline considerably (when done multiple times) due to repeated data transfer between CPU and GPU. 
Most importantly, however, we train our model on batches of CVRP instances which only works effectively due to a shared instance size of all CVRP instances in a batch.
When considering stochasticity during training, the different instances would naturally be of different sizes after partial solution construction, making it much harder to process in existing pipelines.
As a drawback, however, our model cannot learn any hidden patterns of stochasticity that could increase performance during inference such as not visiting certain isolated customers in the beginning due to further expected customers in the same area later on.

\subsection{Making profitable pricing and routing decisions}
\label{sec:cost_delta}

An exact solution of the problem in Section~\ref{sec:operator_price_decision} requires anticipating how accepting a customer affects the cost of serving future, still unknown customers.
GREAT-ESDD instead follows a myopic approach and selects, for each arriving customer, the price menu that maximizes the expected immediate profit.
For an arriving customer in routing state $\tau$, let $\Delta C_j(\tau)$ denote the increase in routing cost if the customer is served via delivery option $j \in \{h, OOH_1, \dots, OOH_K\}$.
Combined with the choice probabilities \eqref{eq:cust_prob}, the expected profit $\pi$ of a price menu $\mathbf p$ is
\begin{equation} \label{eq:exp_profit}
    \mathbb E\bigl[\pi(\mathbf p) \mid \tau, \mathbf d\bigr]
    = \sum_{j \in \{h,\, OOH_1,\dots,OOH_K\}}
        \bigl(p_j - \theta\, \Delta C_j(\tau)\bigr)\,
        \mathbb P(j \mid \mathbf p, \mathbf d),
\end{equation}
where the opt-out option is omitted since its contribution to the profit is exactly zero.
This expected profit depends on the distances $\mathbf d$ of the customer to the OOH locations, which are fixed, on the routing cost increases $\Delta C_j(\tau)$ and on the offered prices $\mathbf p$.
In the following, we describe how we choose the price menu $\mathbf p^\star$ offered to an arriving customer.
Since the prices are taken from the discrete grid $\mathcal G$, we can numerically evaluate
\begin{equation}
    \mathbf p^\star = \arg\max_{\mathbf p \in \mathcal G^{K+1}}
        \mathbb E\bigl[\pi(\mathbf p) \mid \tau, \mathbf d\bigr], \label{eq:argmax_price}
\end{equation}
by enumerating all $|\mathcal G|^{K+1}$ price menus.
For this maximization, it is important to notice that $\Delta C_j(\tau)$ is independent of the offered price.
As a consequence, it suffices to compute this value once per possible delivery location regardless of which price is offered. 
$\Delta C_j(\tau)$ is the part that makes the expected profit, and the steering of the customer's decision, cost aware.
We compute it with the GREAT model from Section~\ref{sec:GREAT}, which gives the GREAT-ESDD framework its name.
Using our trained routing model, we compute the routing cost of the instance with the additional customer's delivery location compared to the prior tour without the additional customer. 
Since our model is capable of batch processing, we can furthermore compute the new route for every different delivery option $j$ for the new customer (different OOH locations and home location) in parallel at low cost.
$\Delta C_j(\tau)$ is then the difference between the cost of the new tour including delivery option $j$ for the new customer and the old tour.
Since both tours are computed heuristically, $\Delta C_j(\tau)$ is an estimate of the marginal routing cost and not the optimal cost increment.
We note that the computation of $\Delta C_j(\tau)$ is therefore completely myopic. 
It does not include any knowledge or intuition about potential future customers who might also still order and by this, potentially make a delivery to a currently isolated customer cheaper. 
We accept this limitation in exchange for the capability to train our GREAT model efficiently and introduce a simple way to mitigate this drawback in the following section.
It is worth noting that optimizing \eqref{eq:argmax_price} requires us to know $\mathbb P(j \mid \mathbf p, \mathbf d)$ in \eqref{eq:exp_profit} which implies that we need to know the customer preference parameters used to compute the utilities. 
While those preferences are generally unknown to a delivery company, we describe in Section~\ref{sec:online_pref} how to learn those parameters online.

\subsection{Rectifying the marginal cost estimate}
\label{sec:rectifying_factor}
As a result of the model's inability to predict future customer arrivals, the estimate $\Delta C_j (\tau)$ and therefore equation \eqref{eq:exp_profit} is biased. 
Consider a customer $i$ who places an order early in the day in a neighborhood of the city where no other customers have ordered yet. 
The estimated increase in routing cost for this customer will consequently be high, since the vehicle is assumed to travel to the neighborhood for customer $i$ only.
However, it is entirely possible that the delivery may take place several hours later and, until then, several further customers order in the same neighborhood. 
Consequently, the marginal cost estimate for the first customer $i$ was inaccurate and too high.
Following this logic, typically, the computed $\Delta C_j(\tau)$ for a delivery option $j$ and routing state $\tau$ will be higher than the true increase in cost. 

Therefore, we additionally introduce a cost correcting rectifier $\omega \in \left[0,1 \right]$ in the computation of the expectation of the profit in the GREAT-ESDD framework:
\begin{equation} \label{eq:prof_scaled}
    \mathbb E\bigl[\pi(\mathbf p) \mid \tau, \mathbf d\bigr]
    = \sum_{j \in \{h,\, OOH_1,\dots,OOH_K\}}
        \bigl(p_j - \omega\, \theta\, \Delta C_j(\tau)\bigr)\,
        \mathbb P(j \mid \mathbf p, \mathbf d),
\end{equation}
where $\omega = 0$ corresponds to a pricing strategy that completely ignores the routing cost and considers the customer preferences only and $\omega = 1$ trusts the raw routing cost estimate.
In our experiments, $\omega$ is a static parameter that we tune offline per routing backend and keep fixed during operations.
The exact value depends on the ``routing backend'' (apart from GREAT, we also test other baselines in our experiments) and is determined by a grid search, see Appendix~\ref{app:omega_search}.

\subsection{Learning unknown customer preferences online}
\label{sec:online_pref}
To compute \eqref{eq:argmax_price} in Section~\ref{sec:cost_delta}, we need to know the probabilities that a customer will accept a delivery option $j$ given prices $\mathbf p$ and distances $\mathbf d$. 
As described in Section~\ref{sec:customer_choice_model}, the probabilities are computed using an MNL model that depends on the utilities of the delivery options for the customers.
These utilities depend on the five unknown parameters $(\alpha_h, \beta_h, \alpha_{OOH}, \beta_{OOH}, \gamma)$.

We model the learning of these parameters as a bandit problem, where every arriving customer corresponds to one round. 
The action of the bandit is offering a price menu $\mathbf p \in \mathcal{G}^{K+1}$, and the context consists of the distance vector $\mathbf d$ and the marginal routing costs $\Delta C_j(\tau)$.
The objective is the expected profit \eqref{eq:prof_scaled}.
Since the profit of an option can be computed once a customer makes a decision, the only unknown components are the parameters of the choice model.
Therefore, we learn from the observed choices $j$ rather than from the realized profits.
We maintain a posterior over the five MNL parameters (note those preferences are shared by all customers) and update it after every customer decision using the likelihood of the observed choice. 
Note that $\mathbf d$ is part of the likelihood, whereas $\Delta C_j(\tau)$ only influences which menu is optimal. 
To balance exploration (offering prices that reduce parameter uncertainty) and exploitation (offering the menu that maximizes the profit \eqref{eq:prof_scaled} under the current estimate), we use Thompson sampling with a Laplace-approximated posterior, described in the following.

First, to avoid constrained optimization, we transform the five utility parameters into an unconstrained space:
\begin{equation}
    \mathbf w = \bigl(\alpha_h,\ \log\beta_h,\ \alpha_{OOH},\ \log\beta_{OOH},\ \log\gamma\bigr)
    \in \mathbb R^5.
\end{equation}
By this we can guarantee that $\beta_h,\beta_{OOH},\gamma > 0$ for any $\mathbf w \in \mathbb R^5$
by exponentiating the corresponding coordinates. 

To initialize our guess for $\mathbf w$ (before any customers have been observed), we place an independent Gaussian prior on each of its coordinates:
\begin{equation}
    \mathbf w \sim \mathcal N(\boldsymbol\mu_0, \boldsymbol\Sigma_0), \qquad
    \boldsymbol\Sigma_0 = \mathrm{diag}\bigl(
        \sigma_\alpha^2,\ \sigma_{\log\beta}^2,\ \sigma_\alpha^2,\
        \sigma_{\log\beta}^2,\ \sigma_{\log\gamma}^2 \bigr).
\end{equation}
Here, the mean $\boldsymbol\mu_0$ encodes an initial domain guess for $(\alpha_h,\log\beta_h,\alpha_{OOH},\log\beta_{OOH},\log\gamma)$ and $\sigma_\alpha,\sigma_{\log\beta},\sigma_{\log\gamma}$ reflect the uncertainty of our guess.

After $t$ customers have entered the system, received a price menu and made a decision, the dataset is defined as
\begin{equation}
    \mathcal D_t = \bigl\{ (\mathbf p^{(n)}, \mathbf d^{(n)}, j^{(n)}) \bigr\}_{n=1}^{t}.
\end{equation}
Here, $\mathbf p^{(n)}$ are the prices offered to customer $n$ and $\mathbf d^{(n)}$ are the distances between the customer and the OOH locations. Further, $j^{(n)} \in \mathcal J$ is the customer's observed choice.

By treating a customer's choice $j^{(n)}$ as conditionally independent of other customers' choices given $\mathbf w$, $\mathbf p^{(n)}$, and $\mathbf d^{(n)}$, the likelihood of the dataset is
\begin{equation}
    \mathcal L(\mathbf w \mid \mathcal D_t)
    = \prod_{n=1}^t \mathbb P\bigl(j^{(n)} \mid \mathbf p^{(n)}, \mathbf d^{(n)}; \mathbf w\bigr).
\end{equation}
Here, $\mathbb P(\cdot)$ is the MNL choice probability defined earlier in \eqref{eq:cust_prob}.

We can now combine the likelihood of the dataset and the prior to obtain the (unnormalized) log-posterior of $\mathbf w$ given all data observed so far:
\begin{equation}
    \log p(\mathbf w \mid \mathcal D_t)
    \ \propto\
    \sum_{n=1}^t \log \mathbb P\bigl(j^{(n)} \mid \mathbf p^{(n)}, \mathbf d^{(n)}; \mathbf w\bigr)
    \ -\ \tfrac12 (\mathbf w - \boldsymbol\mu_0)^\top \boldsymbol\Sigma_0^{-1} (\mathbf w - \boldsymbol\mu_0).
\end{equation}
However, since this posterior is not available in closed form, we approximate it with a \emph{Laplace approximation}. 
First, we compute the maximum a posteriori (MAP) estimate
\begin{equation}
    \mathbf w_t^{\mathrm{MAP}} = \arg\max_{\mathbf w \in \mathbb R^5}
        \log p(\mathbf w \mid \mathcal D_t)
\end{equation}
numerically (using L-BFGS-B; \citealp{Byrd1995}).
We warm-start from the previous iteration's estimate $\mathbf w_{t-1}^{\mathrm{MAP}}$ for efficiency. 
Note that $\mathcal D_t$ is re-used in full each iteration, so the computation is a batch refit rather than a recursive update.

We then approximate the posterior as a Gaussian, where the center is the MAP estimate and the covariance is given by the inverse Hessian of the negative log-posterior \citep{NIPS2011_e53a0a29}:
\begin{equation}
    p(\mathbf w \mid \mathcal D_t) \ \approx\ \mathcal N\bigl(\mathbf w;\ \boldsymbol\mu_t,\ \boldsymbol\Sigma_t\bigr),
    \qquad
    \boldsymbol\mu_t = \mathbf w_t^{\mathrm{MAP}}, \qquad
    \boldsymbol\Sigma_t = \mathbf H_t^{-1},
\end{equation}
where
\begin{equation}
    (\mathbf H_t)_{ik} = -\left.\frac{\partial^2}{\partial w_i\, \partial w_k}
        \log p(\mathbf w \mid \mathcal D_t)\right|_{\mathbf w = \mathbf w_t^{\mathrm{MAP}}}
\end{equation}
is evaluated numerically via central finite differences. 
This numerically computed Hessian is potentially not positive definite, so $\boldsymbol\Sigma_t$ is symmetrized and, in case any eigenvalue is non-positive, shifted by a small multiple of the identity to restore positive definiteness, resulting in a valid covariance matrix.

To trade off exploration and exploitation, we use Thompson sampling.
At the start of iteration $t$ (i.e., when the $t$-th customer orders), before offering any prices, a parameter vector is sampled from the current posterior,
\begin{equation}
    \mathbf w_t \sim \mathcal N(\boldsymbol\mu_{t-1}, \boldsymbol\Sigma_{t-1}).
\end{equation}
This vector is used in place of the true (but unknown) parameters when solving the price optimization problem for that customer. 
After the customer's choice $j^{(t)}$ is observed, the tuple $(\mathbf p^{(t)}, \mathbf d^{(t)}, j^{(t)})$ is added to dataset $\mathcal D_t = \mathcal D_{t-1} \cup \{(\mathbf p^{(t)}, \mathbf d^{(t)}, j^{(t)})\}$.
Then, the posterior is refit as described above to obtain $(\boldsymbol\mu_t, \boldsymbol\Sigma_t)$ to be used in the next iteration.
This Thompson-sampling-based sample-price-observe-refit cycle concentrates the posterior around parameter values that yield near-optimal prices while balancing exploration and exploitation.

Algorithm~\ref{alg:great_esdd} summarizes how the components of Sections~\ref{sec:GREAT}--\ref{sec:online_pref} interact over one operating day. If the customer preferences are known, the sampling and refitting steps are skipped and the true parameters are used instead.

\begin{algorithm}[t]
\caption{GREAT-ESDD over one operating day with online preference learning}
\label{alg:great_esdd}
\begin{algorithmic}[1]
\Require trained GREAT model, rectifier $\omega$, posterior $(\boldsymbol\mu,\boldsymbol\Sigma)$ and data $\mathcal D$ from previous days
\State Initialize the planned route $R$ as a tour over the depot and all OOH locations
\For{each customer $i$ ordering at time $t_i$}
    \State Follow $R$ until $t_i$, update position and load, remove served customers
    \Statex \hspace{\algorithmicindent}(re-solve the residual CVRP whenever a realized demand is lower than ordered)
    \For{each option $j \in \{h, OOH_1,\dots,OOH_K\}$ \textbf{in parallel}}
        \State $R_j \gets$ GREAT solution of the residual CVRP including option $j$ \Comment{Sec.~\ref{sec:GREAT}}
        \State $\Delta C_j \gets C(R_j) - C(R)$ \Comment{Sec.~\ref{sec:cost_delta}}
    \EndFor
    \State Sample $\mathbf w \sim \mathcal N(\boldsymbol\mu, \boldsymbol\Sigma)$ \Comment{Sec.~\ref{sec:online_pref}}
    \State $\mathbf p_i \gets \arg\max_{\mathbf p \in \mathcal G^{K+1}} \sum_{j} \bigl(p_j - \omega\theta\Delta C_j\bigr)\,\mathbb P(j \mid \mathbf p, \mathbf d_i; \mathbf w)$ \Comment{\eqref{eq:argmax_price}, \eqref{eq:prof_scaled}}
    \State Offer $\mathbf p_i$ and observe the choice $j_i \in \mathcal J$
    \If{$j_i \neq \emptyset$}
        \State $R \gets R_{j_i}$
    \EndIf
    \State $\mathcal D \gets \mathcal D \cup \{(\mathbf p_i, \mathbf d_i, j_i)\}$, refit $(\boldsymbol\mu, \boldsymbol\Sigma)$ via MAP and Laplace approximation
\EndFor
\State Serve all remaining stops of $R$ and return to the depot
\end{algorithmic}
\end{algorithm}

\begin{figure}
    \centering
    \includegraphics[width=0.85\textwidth]{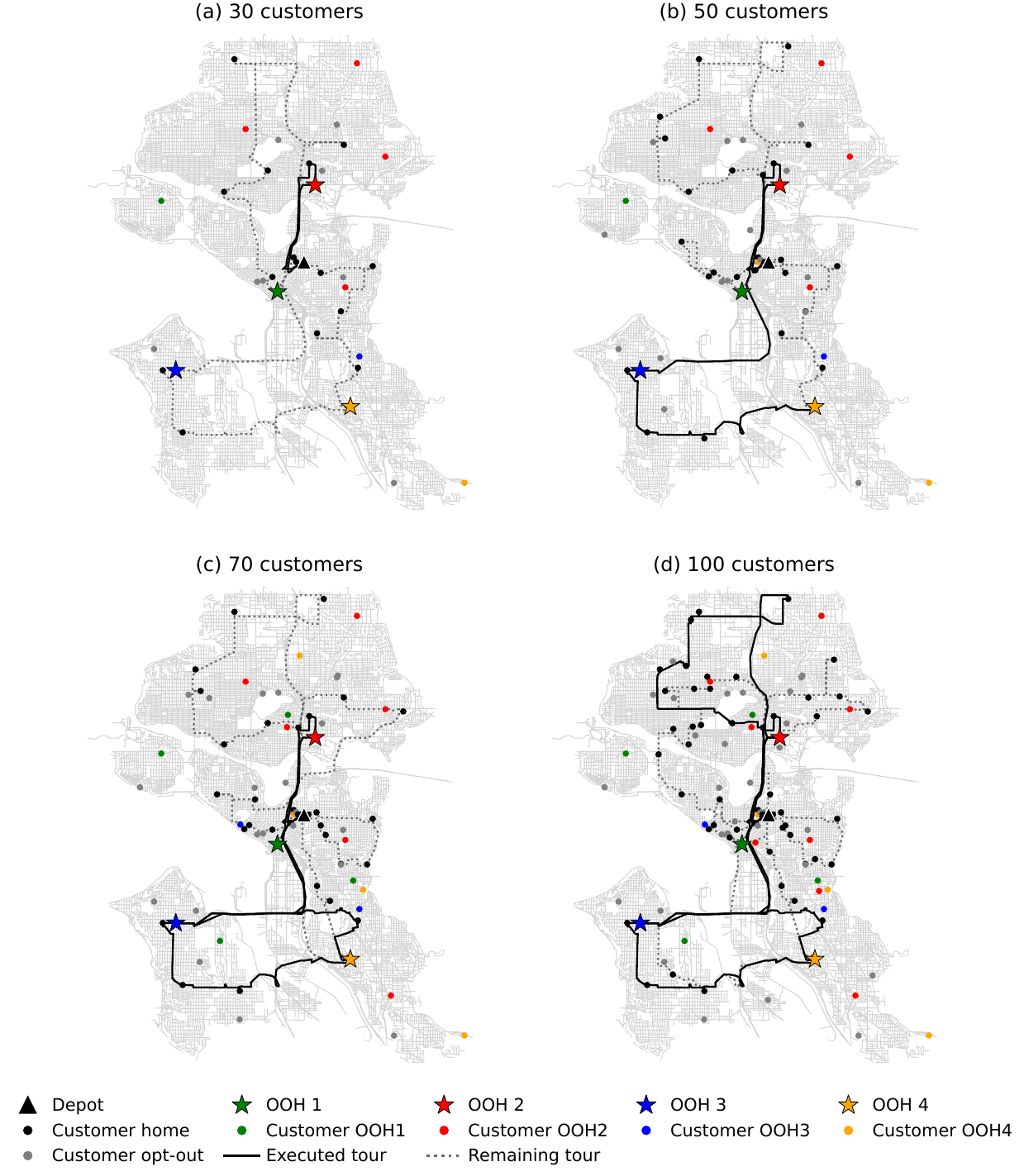}
    \caption{Development of the customer arrival, pricing, customer selection and routing over one day with known customer parameters, shown after (a) 30, (b) 50, (c) 70 and (d) 100 customer requests. The black triangle shows the depot and the colored stars show the OOH locations in our considered Seattle setting. Customers are represented by dots that are either black (home delivery) or colored according to their OOH location; gray dots are customers who opted-out. We show the already executed tour along with the currently still remaining tour.}
    \label{fig:overall}
\end{figure}

\section{Experiments}
\label{sec:experiments}

\subsection{Road network and instance generation}

For our experiments, we consider real-world-derived data. In particular, we use OpenStreetMap \citep{openstreetmap} to obtain a road network graph of Seattle where the nodes correspond to intersections and the edges to their connecting roads. Overall, the graph has $19{,}241$ nodes.
Using the speed limit and distance information of the graph, we derive travel times between all nodes in the graph.
Furthermore, using the Global Human Settlement Population grid \citep{pesaresi2024advances}, we enrich every node in the Seattle graph with its population density.
When sampling customer locations, the nodes are chosen proportionally to this population density, resulting in a realistic setting where our customers are located in areas that also have high population counts in reality. 
We split the graph's nodes into $70\%$ training nodes, $15\%$ validation nodes and $15\%$ testing nodes. For training, customers are only sampled from the training nodes, while for validation and testing customers are only sampled from their respective node sets.
For our pricing-and-routing setting, we further have 5 fixed nodes that correspond to the depot (located in the Capitol Hill district) and four OOH delivery locations, i.e., $K = 4$ (located in Downtown, University District, West and South Seattle).
A visualization can be found in Figure~\ref{fig:overall}.
We note that the distance metric of our data is time-based (representing the travel time between two points in minutes) and it is naturally asymmetric due to one-way streets influencing the travel times.

\subsection{Training the GREAT model}

We train our GREAT-based routing model for 400 epochs. Since we want to train a model capable of solving dynamic instances which naturally have different numbers of active customers over time, we train the model on static instances of sizes $\{10, 20, \dots, 100\}$ with different starting nodes and initial vehicle loads to condition the model to be able to solve partially solved instances. In each epoch, we train the model on $5{,}000$ instances of each instance size. Instances are generated by sampling nodes of the training nodes in our Seattle graph as specified above and using the pairwise travel times between those nodes as our problem defining distances. Every customer node further has a random demand in $\{1, \dots, \bar q\}$ with $\bar q = 9$, and the vehicle has a load capacity of $Q = 50$. Approximately $30\%$ of instances have the depot as the starting node while $70\%$ of instances have any other node as the initial node (uniform probability among those other nodes). 
If the vehicle starts at the depot, it is fully loaded. Otherwise, the amount already delivered is sampled between the start node's demand and $Q - \bar q = 41$, so that the remaining load is at least $\bar q$.
By this, we ensure that every initial move during decoding is valid which helps ensure feasibility during POMO rollouts while conditioning the model to the possibility that other customers have already been served before the current.

Every 10 epochs, a new set of training instances (of all sizes) is sampled. We choose a GREAT model with 5 layers, 128 hidden dimension and 8 attention heads as our encoder, similar to the original GREAT paper. 

Further, we have a validation data set of 200 instances of each size. After every epoch of training we validate our model's current performance and in the end keep the model that performed best out of all 400 epochs of training. 

We choose a learning rate of $0.0001$ and a batch size of $50$ during training. Training was done on a single NVIDIA A40 with 48GB of VRAM.

Training one epoch takes approximately $6.5$ minutes. Further, $45$s are spent on validation every epoch. 
Table~\ref{tab:instance_gen_times} lists the time needed to generate 5,000 new instances of each size (every 10 epochs).

\begin{table}[h]
    \centering
    \begin{tabular}{cc}
        \toprule
        Instance Size & Generation Time \\
        \midrule
        10  & 3s      \\
        20  & 3s      \\
        30  & 6s      \\
        40  & 11s     \\
        50  & 19s     \\
        60  & 31s     \\
        70  & 46s     \\
        80  & 1m 7s   \\
        90  & 1m 20s  \\
        100 & 2m      \\
        \midrule
        \textbf{Sum} & \textbf{6m 26s} \\
        \bottomrule
    \end{tabular}
    \caption{Instance generation times by size}
    \label{tab:instance_gen_times}
\end{table}

The overall training time was therefore approximately $6.5 \cdot 400 + 0.75 \cdot 400 + 6.5 \cdot 40 = 3160$m or roughly 53h.

\subsection{Routing baselines}
In addition to our GREAT-based solver, we also consider other routing backends that can be inserted in a plug-and-play into our pipeline to compute estimates for the marginal routing cost of each delivery option and produce a routing plan for the CVRP instances.

\begin{itemize}
    \item \textbf{GREAT} (ours): we use different versions of our GREAT backend. We evaluate its performance with and without $\times 9$ instance augmentation. Further, we run it on both GPU and CPU to report its runtime on different hardware and consider an additional, very fast, local-search-based postprocessing (details can be found in Appendix~\ref{app:local_search}).
    \item \textbf{PyVRP} \citep{Wouda_Lan_Kool_PyVRP_2024}: a state-of-the-art, open-source VRP solver offering a HGS \citep{vidal2012hybrid} implementation with a Python interface. We run PyVRP with different time limits (up to 1s in the pricing-and-routing setting) resulting in different performance. PyVRP generally runs on a single CPU core which means that we can solve the five candidate CVRPs (arising from the different delivery locations) in parallel on five cores. In the pure CVRP benchmarking of Section~\ref{sec:GREAT_results}, we also run PyVRP without a time limit, which means that we stop the algorithm after no improvement has been found for $20,000$ iterations.
    \item \textbf{Insertion heuristic}: a simple heuristic where at each step every unrouted customer is inserted at the cheapest feasible position in the current tour. The heuristic is very fast and computes the marginal cost at negligible computational cost.
\end{itemize}

\subsection{Pricing simulation setup}

Since real customer responses are unavailable, we simulate customer choices using a ground-truth parameter vector $\mathbf w^\star$ (corresponding to $\alpha_h^\star,\beta_h^\star,\alpha_{OOH}^\star,\beta_{OOH}^\star,\gamma^\star$), which, in the setting of Section~\ref{sec:online_pref}, is \emph{unknown to the learning algorithm}.
We set $\alpha_h^\star=5.0$, $\beta_h^\star=0.6$, $\alpha_{OOH}^\star=4.2$,
$\beta_{OOH}^\star=0.7$ and $\gamma^\star = 0.15$. Using those values, the utility for a \$5 home delivery $U_h(5) = 5 - 0.6 \cdot 5 = 2$ is the same as a \$1 OOH delivery that is 10min away from a customer's home $U_{OOH}(1, 10) = 4.2 - 0.7 \cdot 1 - 0.15 \cdot 10 = 2$. 
To initialize the learnable parameters, we choose the values
\begin{equation}
    \alpha_h = 2.0,\quad \beta_h = 1.0,\quad
    \alpha_{OOH} = 2.0,\quad \beta_{OOH} = 1.0,\quad \gamma = 1.0,
\end{equation} 
as our priors.
We further set
$\sigma_\alpha = 2.0$, $\sigma_{\log\beta} = 1.0$, $\sigma_{\log\gamma} = 1.0$ for the standard deviations of the prior.
Prices are offered from the grid $\mathcal G$ with $p_{\min} = 1$, $p_{\max} = 8$ and $\delta = 0.5$~USD, resulting in $|\mathcal G| = 15$ prices per delivery option and $15^{K+1}$ candidate price menus per customer.

To compute the operation's profit, we assume that one hour of delivery costs the company \$45. Therefore, one minute of delivery has a cost of $\$\tfrac34$. Consequently, we set $\theta = \tfrac34 \frac{\$}{\text{min}}$ to trade off prices paid by the customer and routing cost in equation\eqref{eq:prof_scaled}.
As in training, customer demands are drawn from $\{1, \dots, \bar q\}$ and the vehicle capacity is $Q$.

In our simulation, we assume a random set of 200 customers every day. One fifth of these customers (i.e., 40) order during the ``pre-order'' period during the night, when the vehicle is not yet en route.
The remaining customers arrive dynamically as follows while the vehicle is already operating:  
We solve the CVRP induced by the 40 customers of the pre-order phase using the HGS heuristic \citep{vidal2012hybrid}. Let $L$ be the length (corresponding to travel time) of the resulting CVRP solution.
The $i$-th dynamic customer then arrives at time
\begin{equation}
    t_i = \frac{L}{160}\, i, \qquad i = 1,\dots,160.
\end{equation}
This results in the dynamic arrivals being spread equally over the interval $(0, L]$.

Using this simulation setup, we conduct two types of experiments:
In the first experiment, we assume the customer preference parameters are known and we simulate 25 days of operation of the GREAT-ESDD framework (additionally to other baselines).
The aim of this experiment is to compute certain key performance indicators (KPIs) under different settings.
For example, we are interested in the expected profit the company achieves using different routing backends. 
Further, we are interested in the pricing-decision times that are achieved using those backends. 

Table~\ref{tab:kpis} lists the key performance indicators (KPIs) we compute. They capture the interests of the company, i.e., the profit and how it is achieved through high fees or efficient routing, and the interests of the customers, i.e., how many were served, how much they had to pay and how long they had to wait to obtain prices for their delivery options when ordering.

\begin{table}[h]
\centering
\small
\caption{Key performance indicators of the pricing experiments, averaged over the simulated days.}
\label{tab:kpis}
\begin{tabular}{ll}
\toprule
KPI & Definition \\
\midrule
Profit per day & $\sum_i p_{i,j_i} - \theta C$, see \eqref{eq:daily_profit} \\
Fees per day & $\sum_i p_{i,j_i}$ \\
Routing cost per day & $\theta C$ \\
OOH share & share of customers with $j_i \in \{OOH_1,\dots,OOH_K\}$ \\
Opt-out rate & share of customers with $j_i = \emptyset$ \\
Price paid (mean, median) & $p_{i,j_i}$ for home, OOH and overall, customers with $j_i \neq \emptyset$ \\
Decision time (mean, median, p90, p99) & time to compute all $\Delta C_j(\tau_i)$ and $\mathbf p_i^\star$ \\
\bottomrule
\end{tabular}
\end{table}

We conduct this experiment using two different rectifiers (as introduced in Section~\ref{sec:rectifying_factor}). The first rectifier for each setting is simply $\omega=0$, i.e., we completely ignore routing cost and only take customer preferences into account.
The second rectifier is the best one found for each backend according to Appendix~\ref{app:omega_search}.

As another pricing baseline, we also consider ``flat pricing'' where every delivery option for every customer costs the same. We investigate the best flat price in Appendix~\ref{app:flat_price} which shows that $\$7$ leads to the highest company profit. We use this value whenever flat pricing is mentioned.

In the second type of experiment, we assume that the customer preferences are unknown and need to be learned as outlined in Section~\ref{sec:online_pref}.
In this setting, we simulate 100 days of operations of the GREAT-ESDD framework. 
We track how the customer parameters develop and whether they converge towards the ground-truth values. 
Further, we also track how the offered prices develop with the changing learned parameters and compute the ``profit regret'' when comparing the learned pricing versus the ground truth pricing.

We conduct this experiment twice, once with deterministic customer demands and once with stochastic customer demands as outlined in the lower branch of Figure~\ref{fig:overall_framework}.
To ensure validity of the generated solution, we assume the encountered demand of a customer can only be lower than the originally assumed demand. 
In particular, we assume that $30\%$ of customers randomly have a demand of only $1$ instead of the demand that was planned for.

Moreover, for both the deterministic and stochastic setting, we also run the flat pricing for 100 days and evaluate whether, and if so how quickly, the dynamic pricing can achieve a higher profit than the flat pricing.

\section{Results}
\label{sec:results}

In this section, we start by reporting the raw performance of the trained GREAT routing model in Section~\ref{sec:GREAT_results}. 
Then, in Section~\ref{sec:cost_delta_res}, we move on to reporting the performance and statistics of the pricing-and-routing framework under known customer preferences. Finally, Section~\ref{sec:learning_cust_params_res} reports how well the customer preferences are learned online.

\subsection{GREAT-based CVRP solver}
\label{sec:GREAT_results}
We investigate the performance of our trained GREAT model on CVRP instances of different sizes.
Measuring and quantifying the performance of our neural model is the basis of further experiments where we need fast and reliable performance.

\begin{figure*}
\centering
\includegraphics[width=0.85\textwidth]{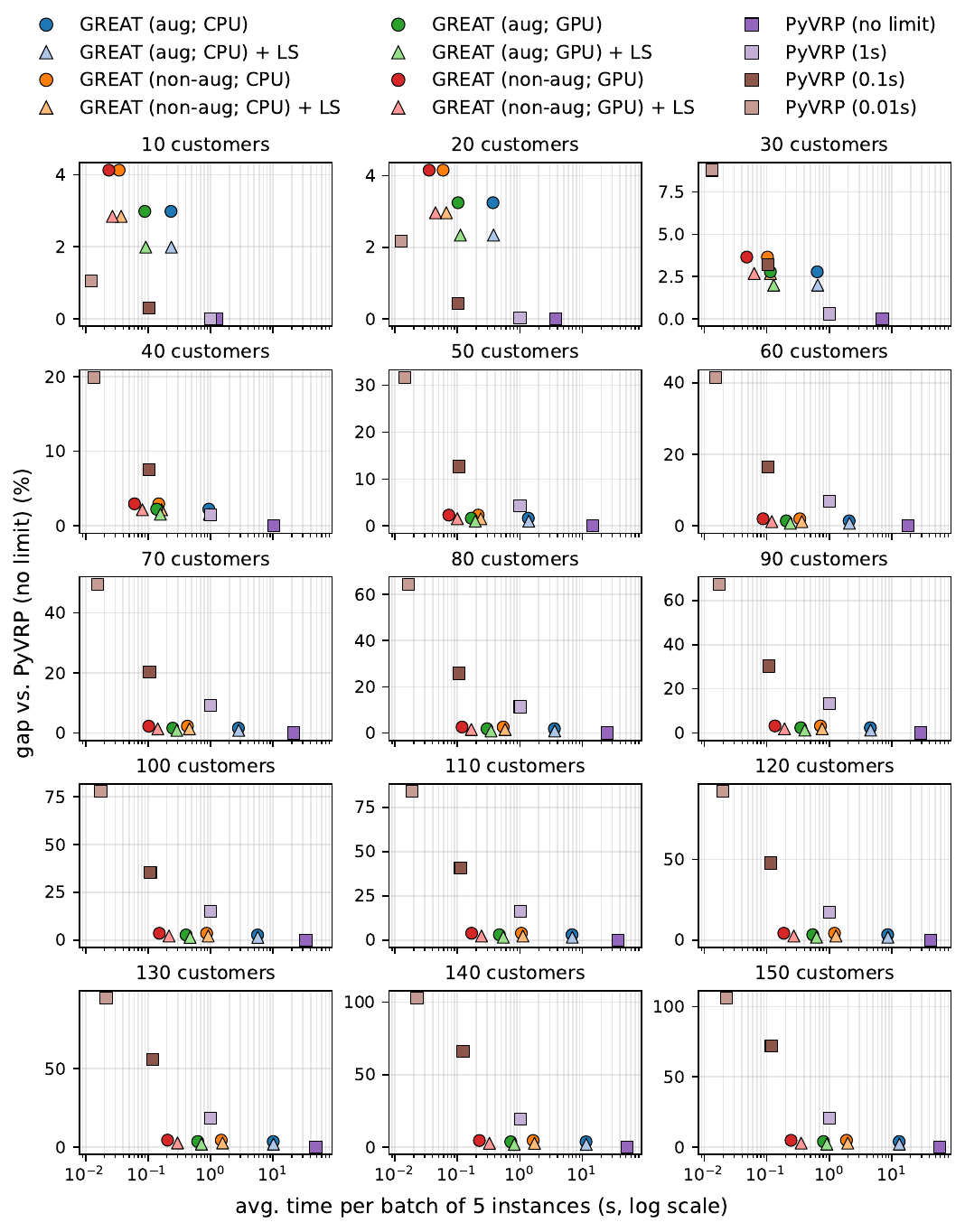}
\caption{Pareto sets trading off performance (gap w.r.t. PyVRP without time limit) vs. runtime for different CVRP instance sizes.}
\label{fig:cvrp_solver_pareto_sets}
\end{figure*}

\begin{figure*}
\centering
\includegraphics[width=\textwidth]{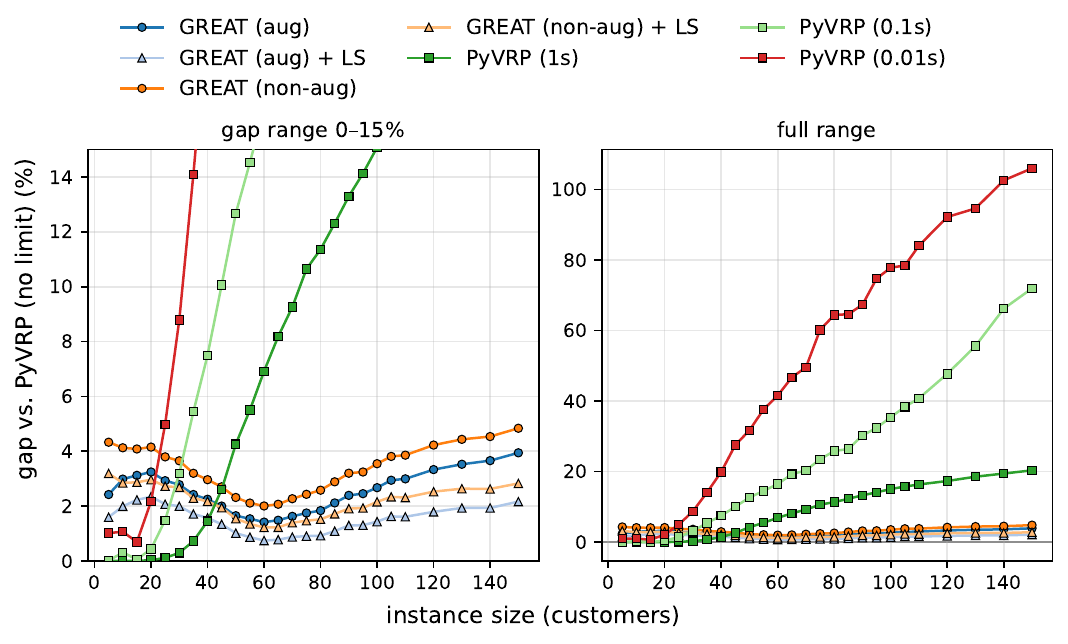}
\caption{Performance of different CVRP solvers w.r.t. PyVRP without time limit. Left: gap range 0--15\,\%; right: unconstrained range.}
\label{fig:cvrp_solver_performance}
\end{figure*}

\begin{figure*}
\centering
\includegraphics[width=\textwidth]{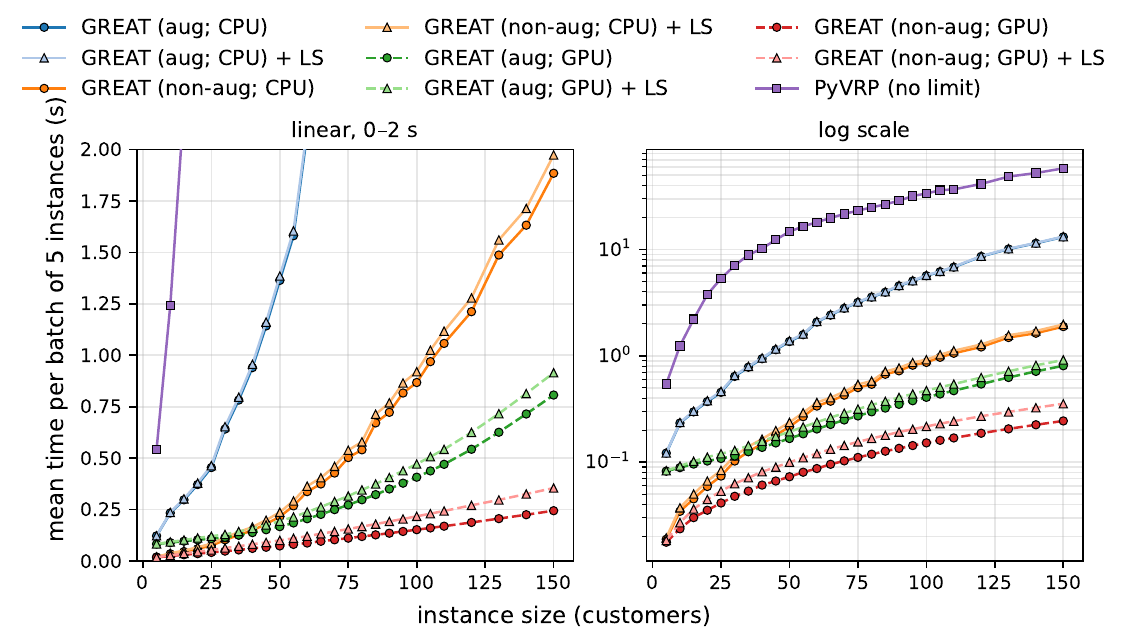}
\caption{Runtime of different CVRP solvers (PyVRP with set time limit is omitted) to solve a batch of 5 CVRP instances in parallel. Left: linear scale, 0--2\,s; right: logarithmic scale, unconstrained range.}
\label{fig:cvrp_solver_speed}
\end{figure*}

Figures~\ref{fig:cvrp_solver_pareto_sets}, \ref{fig:cvrp_solver_performance} and \ref{fig:cvrp_solver_speed} show the performance and runtime of the neural solver based on hardware (CPU vs. GPU), augmentation factor (augmented or non-augmented) and with and without local search. Further, we show the results of several runs of PyVRP with different time limits (0.01s, 0.1s, 1s and no time limit per instance). We show plots for instances of sizes $10, 20, \dots, 150$ and note that instances of sizes bigger than 100 are out of training distribution. The reported performance is the average gap over 1000 test instances of each size. The runtimes are the wall clock times it takes on average to solve a batch of 5 instances (we choose a batch size of 5 since in our pricing framework we also need to solve 5 instances in parallel). For PyVRP, the 5 instances run in parallel on different CPU cores, whereas for the neural model the 5 instances are processed in parallel as PyTorch batches.

The results show that GREAT naturally gets slower with increased instance size since GREAT does not run on a fixed time budget. 
As expected, GREAT is much faster when running on GPUs. Further, skipping augmentation can reduce runtime while not losing too much performance. Instead, the local search postprocessing can be used as an alternative, more effective and efficient way to improve GREAT solution quality, as we can see that no augmentation with postprocessing generally achieves faster runtime and better performance than augmentation without postprocessing.

Runtime-wise, we can see that GREAT on GPUs can achieve a runtime of 0.1s for 70 customers (without augmentation and without postprocessing) while achieving a much better performance than PyVRP under this time budget. 
For 45 customers or more, GREAT achieves better performance than PyVRP with 1s budget. For a time budget of 0.1s, GREAT is generally competitive with PyVRP on instances of size 30. For 20 customers and below, GREAT generally can not compete with PyVRP.

Figure~\ref{fig:cvrp_solver_performance} also underlines that GREAT overtakes PyVRP in terms of performance for 40-50 customers (depending on augmentation factor and postprocessing), whereas for instances with 20 customers or less, GREAT is generally not competitive. 
We point out that the figure shows additional results for instances with $5, 15, \dots, 105$ customers. We can see that the performance on these out-of-training-distribution instances is generally not much different than on equally big in-distribution instances, which confirms that the model generalizes. The plot shows GREAT performs best on instances with approximately 60 customers, which is roughly the average of the training instance sizes ($\frac{1}{10}\sum_{i=1}^{10} 10i = 55$). We can see that generalization performance of GREAT to larger instances is good, with gaps of below $5\%$ on instances of 150 customers with neither augmentation nor postprocessing and approximately $2\%$ with both augmentation and postprocessing.
More importantly, the gap of GREAT stays within a narrow band of approximately 1--5\% over all instance sizes from 5 to 150, whereas the gap of PyVRP under a fixed time limit grows steadily with the instance size, e.g., to approximately 20\% for 150 customers with a 1s limit. This stability across instance sizes is what makes GREAT suitable for our online setting, in which the number of active customers fluctuates over the day.

While the computational complexity of our neural solver grows quadratically with $n$, the runtime increases only slightly for the considered instance sizes when running on GPUs. We can see that when running on GPUs, even for instances of size 150, the time required to solve a batch of 5 instances stays below 1s. When running on a CPU, we can achieve a runtime of under 1s for instances of size 100 and a time of under 2s for instances of size 150 when skipping augmentation.

\subsection{Pricing with known customer preferences} \label{sec:cost_delta_res}

Before we try to learn the pricing parameters, we compare routing backends to see how they perform under a setting of complete knowledge, i.e., with known parameters for the customer utility computations.

For this comparison, we test different routing algorithms. First, we test the neural GREAT approach with and without augmentation on GPU and CPU devices (resulting in different runtimes and performance). 
Since we have seen in Section~\ref{sec:GREAT_results} that local search is a cheap way to consistently improve performance, all GREAT-based routing models use this postprocessing method.
Further, we run PyVRP with different time limits (0.01s, 0.1s, 0.2s, 0.5s and 1s to make a decision) as well as a simple, very fast insertion heuristic that computes routes at negligible computational cost but with lower solution quality. 
Since those baselines run on a single CPU core, we can effectively parallelize them to solve the 5 different CVRPs for the different delivery options in parallel when a customer orders.
For all these options, we try two rectifier values: First, we set $\omega = 0$, which results in a pricing strategy that ignores the delivery cost and only optimizes for the customers preferences. Second, we set $\omega$ to the best value that we empirically found in the discrete grid of $\{0, 0.1, 0.2, \dots, 1\}$.
Details regarding the search for the optimal $\omega$ (which can be different for each routing algorithm) can be found in Appendix~\ref{app:omega_search}.

\begin{figure}
    \centering
    \includegraphics[width=0.5\linewidth]{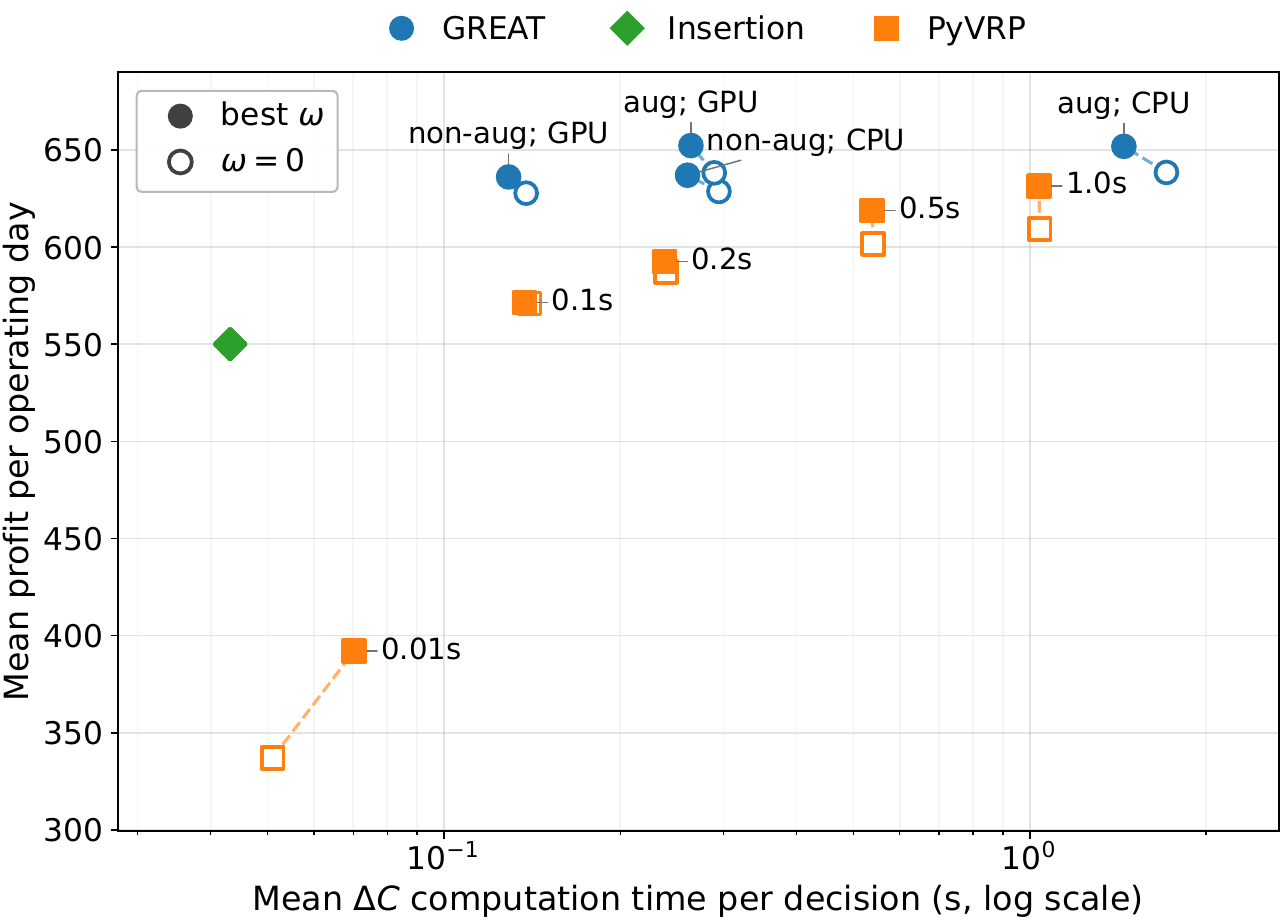}
    \caption{Marginal routing cost ablation, showing mean time to compute the routing cost increase for a pricing decision vs. the mean profit per operating day over 25 days}
    \label{fig:cost_delta}
\end{figure}

A figure showing the mean performance of the different routing backends with the best rectifier $\omega$ and $\omega=0$ (i.e., no routing cost) can be observed in Figure~\ref{fig:cost_delta}.
We can see that when running on a GPU (the same NVIDIA A40 used for training), the mean runtime of the GREAT-ESDD framework without augmentation coincides with the mean runtime of PyVRP with 0.1s limit. 
However, we can also see that average profit per day using GREAT-ESDD is approximately $\$65$, or 11.3\%, higher. 
Similarly, with augmentation, GREAT-ESDD takes approximately 0.2s per decision. 
The comparable PyVRP backend with a 0.2s limit achieves approximately $\$60$ per day profit less.
We further note that when running on a CPU cluster with 32 cores (the same cluster that was used for the PyVRP results), GREAT-ESDD takes approximately 0.3s without and 1.6s with augmentation per pricing decision.
We can see that GREAT-ESDD generally performs better than PyVRP under these time constraints, with PyVRP with 1s time limit performing about as well as GREAT without augmentation. 
A further observation is that in the majority of cases considering the $\omega$ rectified routing costs leads to an improvement. 
The only exception to this is the insertion heuristic (for which the best rectifier was found to be 0).
The sensitivity of the profit to $\omega$ for all backends is shown in Appendix~\ref{app:omega_search}.

Precise numbers for the results in Figure~\ref{fig:cost_delta} can be found in Tables~\ref{tab:econ}, \ref{tab:service} and \ref{tab:prices_times}.
Table~\ref{tab:econ} focuses on the company's profit, whereas Table~\ref{tab:service} shows customer statistics (OOH share and customer opt-out rate) and Table~\ref{tab:prices_times} shows further customer statistics (price paid and waiting times to receive a price).
Note that the times in Table~\ref{tab:prices_times} include the pricing maximization times of equation \eqref{eq:argmax_price} in addition to the marginal routing cost computation times that are reported in Figure~\ref{fig:cost_delta}. 
The pricing maximization adds a constant overhead of approximately $0.2$s (observable in the decision time entry for the insertion heuristic) to all pricing decisions on top of the marginal routing cost computation.
Results of GPU and CPU runs of GREAT-ESDD are statistically equivalent (in terms of performance), the small differences stem merely from inconsistencies of PyTorch implementations for different devices\footnote{https://docs.pytorch.org/docs/2.14/notes/randomness.html}. 
Table~\ref{tab:econ} also contains 95\% confidence intervals comparing paired difference in mean profit against two baselines: the dynamic PyVRP with 0.5s time limit and the flat pricing. 
Since the 25 simulated days use the identical instances and seeding per backend, every simulation is a genuine matched pair.

When comparing the performance of the dynamic pricing with the flat pricing strategy, we can see that the flat pricing (which uses augmented GREAT for routing) achieves approximately the performance of the dynamic pricing of the non-augmented GREAT backend in terms of profit. 
When compared to the same augmented backend, the profit is on average around $\$14$ per day lower.
Although flat pricing collects more fees, the considerably higher routing cost (due to lower OOH share) leads to lower profit.
Further, the opt-out rate is slightly higher than for dynamic pricing based runs. 
Table~\ref{tab:prices_times} shows that the augmented GREAT-ESDD results in an average overall price of $\$6.78$ compared to the \$7 for the flat pricing, a $3.2\%$ discount. 
OOH deliveries are on average $\$6.07$ and home deliveries $\$7.07$, a $\sim13.3\%$ discount and $1\%$ increase, respectively, relative to flat pricing.

Overall, we can summarize that the GREAT-ESDD-based dynamic pricing outperforms the PyVRP-based dynamic pricing substantially, especially under similar runtime budgets.
Moreover, dynamic pricing steers more customers to OOH delivery and results in lower prices paid by customers, while even slightly increasing the company's profit.
Both results combined show that it is important to effectively steer the customer decisions, as neither flat pricing (with high collected fees but low OOH quota) nor PyVRP with low time limits (which has high OOH quota but low collected fees) perform better than GREAT-ESDD.

\begin{table*}[!t]
\centering
\small
\caption{Economic outcome of the pricing-and-routing operation, averaged over 25 simulated days. Each routing backend uses its best rectifier $\omega$. Profit is fees minus routing cost. Best value per column in bold. $\Delta$ is the paired difference in mean profit against a baseline.}

\label{tab:econ}
\begin{tabular}{llrrrrrr}
\toprule
 & & & \multicolumn{5}{c}{Mean per day (\$)} \\
\cmidrule(lr){4-8}
Backend & \shortstack{Device /\\limit} & $\omega$ & Profit & \shortstack{$\Delta$ vs.\ PyVRP 0.5\,s\\(95\% CI)} & \shortstack{$\Delta$ vs.\ flat\\(95\% CI)} & Fees & Routing cost \\
\midrule
GREAT (non-aug) & GPU & 0.40 & 636.18 & +17.37 $\pm$ 10.77 & -2.20 $\pm$ 8.92 & 1008.06 & \textbf{371.88} \\
GREAT (aug) & GPU & 0.30 & \textbf{652.35} & +33.54 $\pm$ 10.01 & +13.97 $\pm$ 9.51 & 1027.94 & 375.59 \\
GREAT (non-aug) & CPU & 0.40 & 637.08 & +18.26 $\pm$ 10.66 & -1.31 $\pm$ 8.73 & 1009.66 & 372.58 \\
GREAT (aug) & CPU & 0.30 & 651.95 & +33.14 $\pm$ 9.79 & +13.57 $\pm$ 8.97 & 1028.66 & 376.71 \\
\midrule
PyVRP & 0.01\,s & 0.10 & 392.07 & -226.75 $\pm$ 22.89 & -246.32 $\pm$ 22.14 & 937.16 & 545.09 \\
PyVRP & 0.1\,s & 0.30 & 571.48 & -47.34 $\pm$ 13.80 & -66.91 $\pm$ 12.91 & 977.86 & 406.38 \\
PyVRP & 0.2\,s & 0.20 & 592.68 & -26.14 $\pm$ 9.84 & -45.71 $\pm$ 10.00 & 1016.04 & 423.36 \\
PyVRP & 0.5\,s & 0.30 & 618.82 & (ref.) & -19.57 $\pm$ 8.65 & 1010.82 & 392.00 \\
PyVRP & 1\,s & 0.30 & 631.56 & +12.74 $\pm$ 7.03 & -6.83 $\pm$ 7.39 & 1018.32 & 386.76 \\
\midrule
Insertion heuristic & -- & 0.00 & 550.09 & -68.73 $\pm$ 12.44 & -88.30 $\pm$ 11.01 & \textbf{1043.06} & 492.97 \\
Flat pricing (\$7) & -- & -- & 638.39 & +19.57 $\pm$ 8.65 & (ref.) & 1038.24 & 399.85 \\
\bottomrule
\end{tabular}
\end{table*}

\begin{table*}[!t]
\centering
\small
\caption{Service outcome from the customers' perspective, averaged over 25 simulated days. Each routing backend uses its best rectifier $\omega$.}
\label{tab:service}
\begin{tabular}{llrr}
\toprule
Backend & \shortstack{Device /\\limit} & OOH share & Opt-out rate \\
\midrule
GREAT (non-aug) & GPU & 33.3\% & 24.54\% \\
GREAT (aug) & GPU & 29.2\% & 24.18\% \\
GREAT (non-aug) & CPU & 33.4\% & 24.44\% \\
GREAT (aug) & CPU & 29.3\% & 24.12\% \\
\midrule
PyVRP & 0.01\,s & 36.8\% & 21.24\% \\
PyVRP & 0.1\,s & 34.8\% & 24.12\% \\
PyVRP & 0.2\,s & 28.9\% & 24.20\% \\
PyVRP & 0.5\,s & 30.8\% & 24.84\% \\
PyVRP & 1\,s & 30.9\% & 24.52\% \\
\midrule
Insertion heuristic & -- & 23.8\% & 22.84\% \\
Flat pricing (\$7) & -- & 18.6\% & 25.84\% \\
\bottomrule
\end{tabular}
\end{table*}

\begin{table*}[!t]
\centering
\small
\caption{Realized prices paid by customers and the distribution of the price-decision time (marginal routing cost computation plus price selection), over 25 simulated days. Each routing backend uses its best rectifier $\omega$.}
\label{tab:prices_times}
\begin{tabular}{llrrrrrrrrrr}
\toprule
 & & \multicolumn{6}{c}{Price paid (\$)} & \multicolumn{4}{c}{Decision time (s)} \\
\cmidrule(lr){3-8}\cmidrule(lr){9-12}
 & & \multicolumn{2}{c}{Home} & \multicolumn{2}{c}{OOH} & \multicolumn{2}{c}{Overall} & & & & \\
Backend & \shortstack{Device /\\limit} & Mean & Med. & Mean & Med. & Mean & Med. & Mean & Med. & p90 & p99 \\
\midrule
GREAT (non-aug) & GPU & 7.14 & 7.00 & 5.75 & 6.00 & 6.68 & 7.00 & 0.337 & 0.333 & 0.424 & 0.472 \\
GREAT (aug) & GPU & 7.07 & 7.00 & 6.07 & 6.00 & 6.78 & 7.00 & 0.472 & 0.443 & 0.676 & 0.797 \\
GREAT (non-aug) & CPU & 7.15 & 7.00 & 5.75 & 6.00 & 6.68 & 7.00 & 0.417 & 0.381 & 0.689 & 0.844 \\
GREAT (aug) & CPU & 7.07 & 7.00 & 6.07 & 6.00 & 6.78 & 7.00 & 1.604 & 1.336 & 3.395 & 4.967 \\
\midrule
PyVRP & 0.01\,s & 6.93 & 7.00 & 4.27 & 4.50 & 5.95 & 6.50 & 0.227 & 0.208 & 0.228 & 0.261 \\
PyVRP & 0.1\,s & 7.13 & 7.00 & 5.16 & 5.50 & 6.44 & 7.00 & 0.294 & 0.294 & 0.309 & 0.321 \\
PyVRP & 0.2\,s & 7.04 & 7.00 & 5.88 & 6.00 & 6.70 & 7.00 & 0.394 & 0.396 & 0.410 & 0.420 \\
PyVRP & 0.5\,s & 7.12 & 7.00 & 5.84 & 6.00 & 6.72 & 7.00 & 0.695 & 0.695 & 0.710 & 0.721 \\
PyVRP & 1\,s & 7.11 & 7.00 & 5.94 & 6.00 & 6.75 & 7.00 & 1.193 & 1.196 & 1.211 & 1.222 \\
\midrule
Insertion heuristic & -- & 6.84 & 7.00 & 6.51 & 6.50 & 6.76 & 7.00 & 0.200 & 0.199 & 0.229 & 0.243 \\
Flat pricing (\$7) & -- & 7.00 & 7.00 & 7.00 & 7.00 & 7.00 & 7.00 & -- & -- & -- & -- \\
\bottomrule
\end{tabular}
\end{table*}

\subsection{Learning to price} \label{sec:learning_cust_params_res}

In this section, we now assume that the customer utility parameters are unknown. 
The goal of the experiment is to try to recover those unknown parameters, in a deterministic (customers have known demands) and stochastic (customer demands can be lower than expected) setting.
Since we already know from Section~\ref{sec:cost_delta_res} that the augmented GREAT model achieves the highest profits, we use this routing model as our backend.

\begin{figure*}
\centering
\begin{subfigure}[b]{0.32\textwidth}
    \centering
    \includegraphics[width=\textwidth]{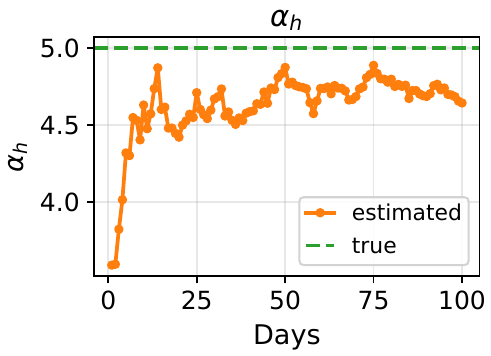}
\end{subfigure}
\hfill
\begin{subfigure}[b]{0.32\textwidth}
    \centering
    \includegraphics[width=\textwidth]{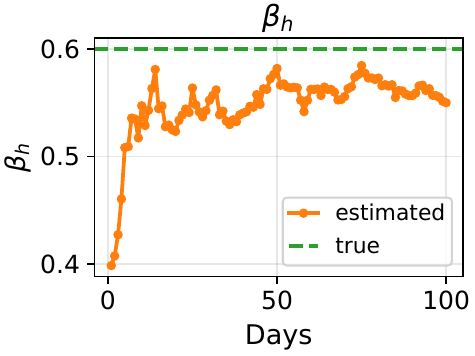}
\end{subfigure}
\hfill
\begin{subfigure}[b]{0.32\textwidth}
    \centering
    \includegraphics[width=\textwidth]{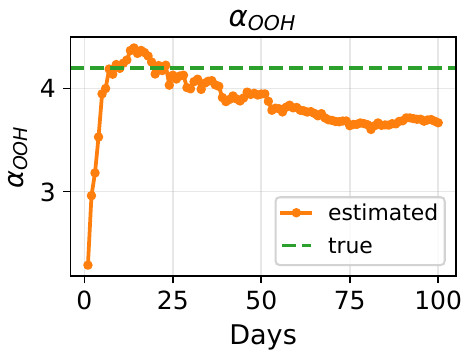}
\end{subfigure}

\vspace{0.6em}

\begin{subfigure}[b]{0.32\textwidth}
    \centering
    \includegraphics[width=\textwidth]{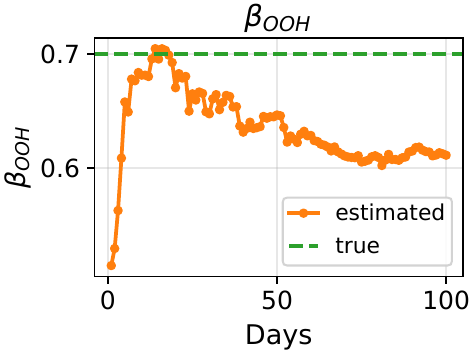}
\end{subfigure}
\hfill
\begin{subfigure}[b]{0.32\textwidth}
    \centering
    \includegraphics[width=\textwidth]{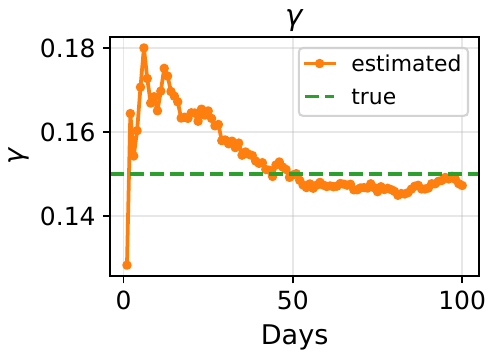}
\end{subfigure}
\hfill
\begin{subfigure}[b]{0.32\textwidth}
    \centering
    \includegraphics[width=\textwidth]{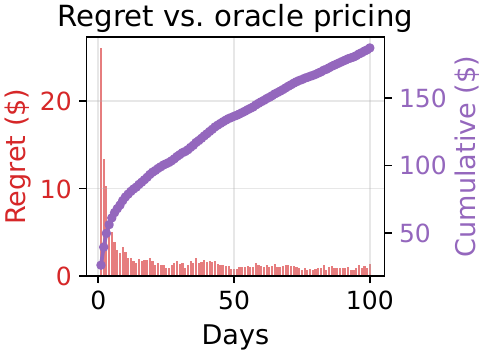}
\end{subfigure}

\vspace{0.6em}

\begin{subfigure}[b]{0.32\textwidth}
    \centering
    \includegraphics[width=\textwidth]{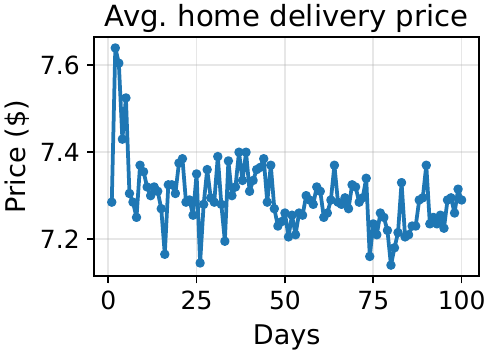}
\end{subfigure}
\hfill
\begin{subfigure}[b]{0.32\textwidth}
    \centering
    \includegraphics[width=\textwidth]{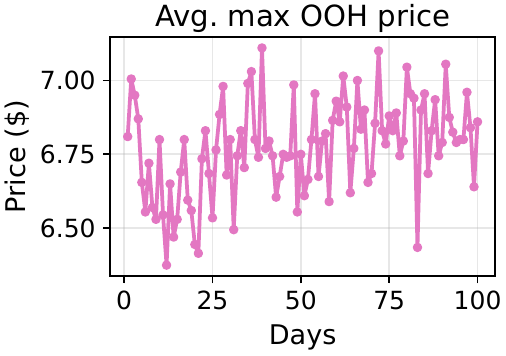}
\end{subfigure}
\hfill
\begin{subfigure}[b]{0.32\textwidth}
    \centering
    \includegraphics[width=\textwidth]{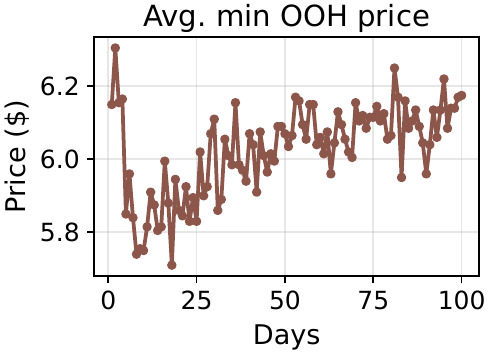}
\end{subfigure}

\caption{Parameter estimation, price and regret developments of the deterministic learning to price setting.}
\label{fig:deterministic_leaning_to_price}
\end{figure*}

\begin{figure*}
\centering
\begin{subfigure}[b]{0.32\textwidth}
    \centering
    \includegraphics[width=\textwidth]{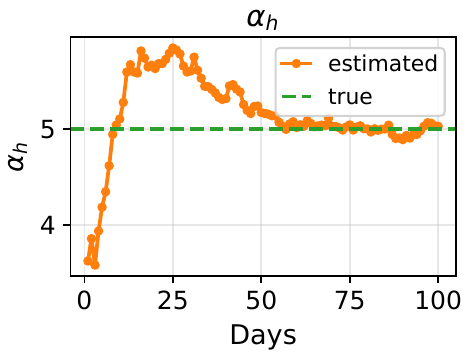}
\end{subfigure}
\hfill
\begin{subfigure}[b]{0.32\textwidth}
    \centering
    \includegraphics[width=\textwidth]{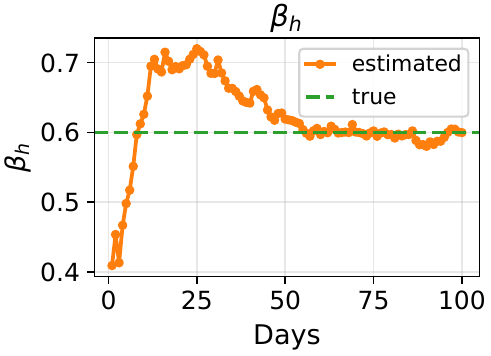}
\end{subfigure}
\hfill
\begin{subfigure}[b]{0.32\textwidth}
    \centering
    \includegraphics[width=\textwidth]{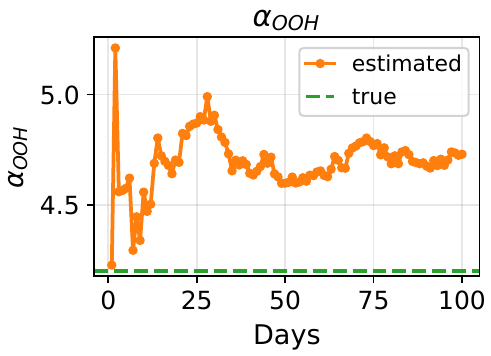}
\end{subfigure}

\vspace{0.6em}

\begin{subfigure}[b]{0.32\textwidth}
    \centering
    \includegraphics[width=\textwidth]{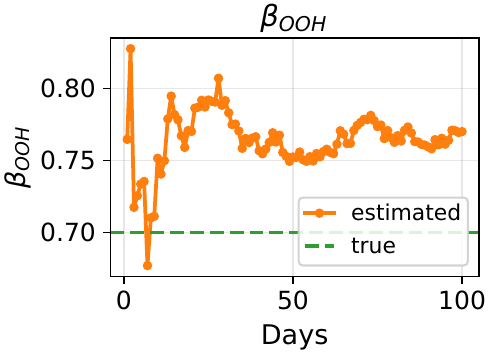}
\end{subfigure}
\hfill
\begin{subfigure}[b]{0.32\textwidth}
    \centering
    \includegraphics[width=\textwidth]{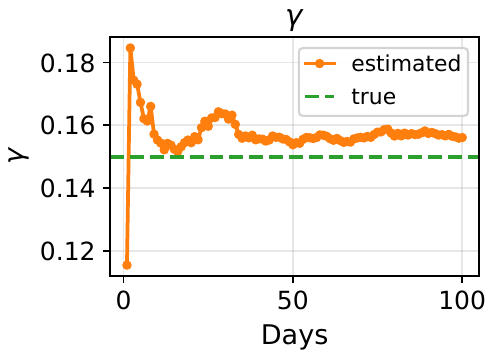}
\end{subfigure}
\hfill
\begin{subfigure}[b]{0.32\textwidth}
    \centering
    \includegraphics[width=\textwidth]{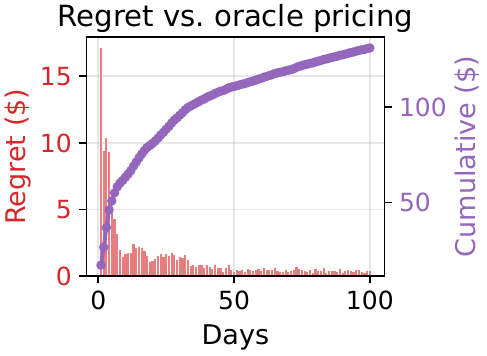}
\end{subfigure}

\vspace{0.6em}

\begin{subfigure}[b]{0.32\textwidth}
    \centering
    \includegraphics[width=\textwidth]{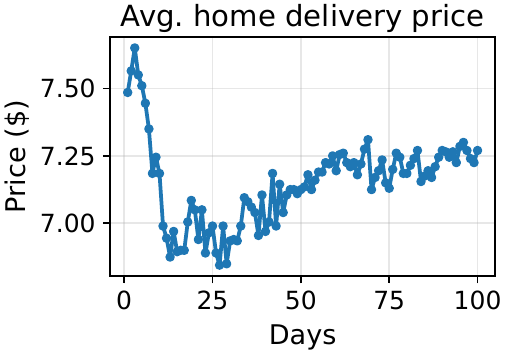}
\end{subfigure}
\hfill
\begin{subfigure}[b]{0.32\textwidth}
    \centering
    \includegraphics[width=\textwidth]{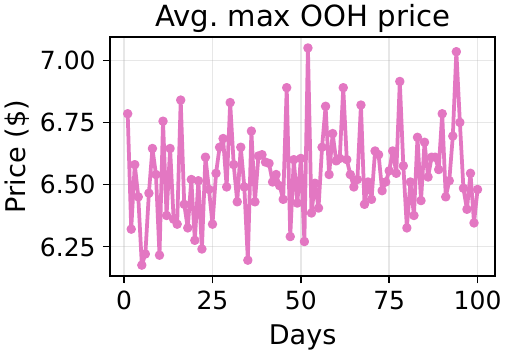}
\end{subfigure}
\hfill
\begin{subfigure}[b]{0.32\textwidth}
    \centering
    \includegraphics[width=\textwidth]{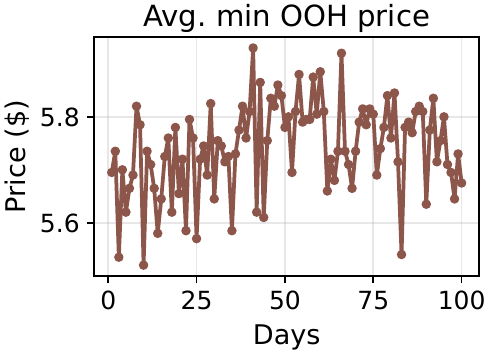}
\end{subfigure}

\caption{Parameter estimation, price and regret developments of the stochastic learning to price setting.}
\label{fig:stochastic_leaning_to_price}
\end{figure*}

The results of the experiment can be seen in Figure~\ref{fig:deterministic_leaning_to_price} (deterministic setting) and Figure~\ref{fig:stochastic_leaning_to_price} (stochastic setting). 
The figures show that the parameters are approximately recovered and the values seem to converge after around 50 days of simulation. 
In the deterministic setting, the learned parameters tend to be lower than the unknown ground-truth values. 
However, all parameters converge to a value in the correct order of magnitude. 
In the stochastic setting, we observe similar results with $\alpha_h, \beta_h$ and $\gamma$ being recovered almost perfectly. $\alpha_{OOH}$ and $\beta_{OOH}$ are slightly overestimated, however. 
We attribute the difference in parameter recovery mainly to seed noise, since the stochastic demands influence the Thompson sampling only indirectly through the marginal routing costs and, consequently, through the offered price menus.
The cumulative regret (comparing expected profit from the ground-truth oracle vs. the actual profit) for both settings grows sublinearly, as expected for Thompson sampling, indicating that learning is effective and underlining that low regret does not require exact recovery.
In the deterministic setting, $\alpha$ and $\beta$ are underestimated jointly, so that their ratio, which determines the price at which an option becomes unattractive, is recovered closely for both home and OOH delivery.

The bottom row of the plots show the average home delivery prices, average price of the most expensive offered OOH option and cheapest offered OOH option over all customers of one simulation day.
While the home delivery prices seem to change over the course of the learning and converge, the maximum OOH prices are mostly noise. 
Similarly, the minimum OOH prices seem to be noise in the stochastic setting, however, in the deterministic setting we can see that after an initial drop, there seems to be a trend of increasing minimum prices.

\begin{figure*}
\centering
\begin{subfigure}[b]{0.48\textwidth}
    \centering
    \includegraphics[width=\textwidth]{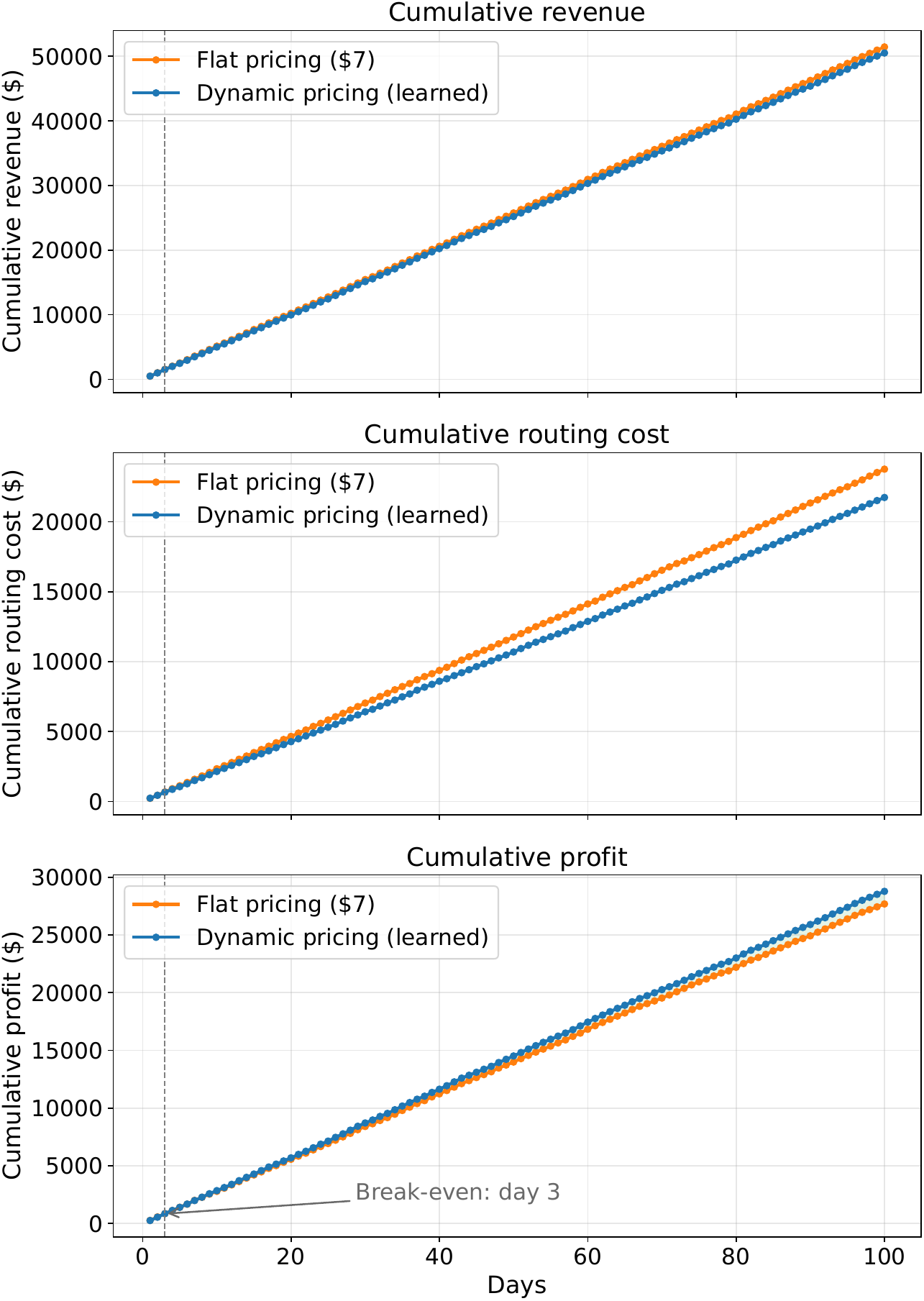}
    \caption{Deterministic Setting}
\end{subfigure}
\hfill
\begin{subfigure}[b]{0.48\textwidth}
    \centering
    \includegraphics[width=\textwidth]{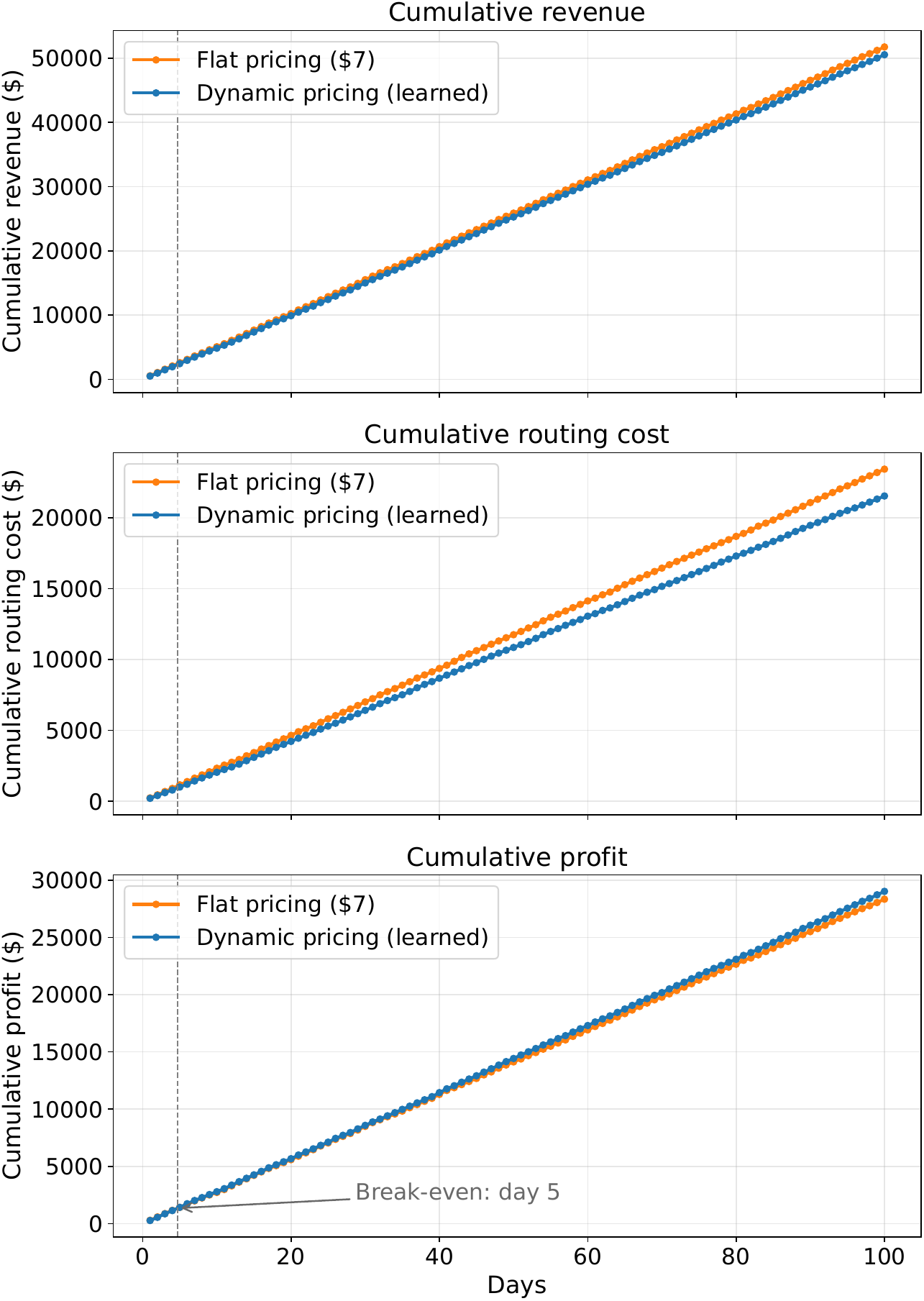}
    \caption{Stochastic Setting}
\end{subfigure}
\caption{Cumulative revenue, routing cost and profit over 100 days of operation, using the bandit-based pricing and the flat \$7 pricing strategies in the deterministic and stochastic setting.}
\label{fig:comparison_bandit_flat}
\end{figure*}

For comparison with flat $\$7$ pricing, we also simulate the 100 day operation with such a flat strategy. 
The result is shown in Figure~\ref{fig:comparison_bandit_flat}. 
As we know from Section~\ref{sec:cost_delta_res}, the expected difference in profit of the flat pricing and dynamic pricing (with the correct customer parameters) is $\$14$ a day. 
However, since the parameter estimates in the beginning are wrong, it might take several days to make up for a suboptimal pricing in the beginning. 
Figure~\ref{fig:comparison_bandit_flat} shows the break-even point of the dynamic pricing to be after 3-5 days (depending on whether we deal with the deterministic or stochastic setting). 
Moreover, the difference in cumulative profit seems to be on the order of $\$1000$ after 100 days of simulation, which is a similar order of magnitude as the $\$14$ a day reported in Section~\ref{sec:cost_delta_res}.

We conclude that the parameters of the customer preferences can indeed be learned. 
We hypothesize that more extensive learning (more than 100 days) or better-informed priors could lead to better results.
We emphasize that even under unknown customer parameters, we can achieve better results than the flat pricing within $5$ days.

\section{Discussion of limitations} \label{sec:limitations}
As is, GREAT-ESDD has the following limitations.

\paragraph{Myopic $\Delta C$ estimate.} As outlined in Section~\ref{sec:cost_delta}, the marginal routing cost estimate is myopic, since our solver is trained on deterministic data and cannot anticipate future customer arrivals. As a consequence, early or isolated orders tend to be overpriced, which distorts the steering of customer decisions. A model that anticipates future arrivals could produce more strategic routes (e.g., visit isolated customers only later when more customers in the same neighborhood are expected) and better estimates. However, it would require modeling time and customer arrivals during training, lose the training efficiency discussed in Sections~\ref{sec:dynamic_routing_literature} and~\ref{sec:GREAT}, and be conditioned on a specific city and customer distribution. Nevertheless, when a specific customer distribution and city is fixed, such a distribution-aware model would potentially result in better performance of GREAT-ESDD.

\paragraph{Heuristic marginal cost.} $\Delta C_j(\tau)$ is the difference between two approximate route costs. It is therefore not the optimal cost increment, and the errors of both solutions carry over to the estimate. Our results indicate that solution quality matters, as backends with smaller gaps generally achieve higher profits (Section~\ref{sec:cost_delta_res}), but we do not quantify the estimation error directly.

\paragraph{Static rectifier $\omega$.} GREAT-ESDD uses a fixed rectifier $\omega$ that is tuned offline per routing backend in a simple grid search. While this works (including the rectified marginal routing cost in the optimal price search produces better results than both non-rectified $\Delta C$ and no routing cost at all), it is suboptimal. Since $\omega$ absorbs the expected cost savings from future customers until the end of the day, its value depends on the city, the demand pattern and the operating horizon, and it may not transfer to other settings without re-tuning. Our results suggest that $\omega$ should be tuned per backend on a few simulated days of the target setting and that noisier cost estimates call for smaller values (Appendix~\ref{app:omega_search}). We hypothesize that an $\omega$ that changes as a function of time, or that is tuned online, could result in better performance, since the $\Delta C$ estimates become more accurate over the day. In fact, the estimate of the very last customer who orders is no longer affected by future arrivals at all.

\paragraph{Single vehicle CVRP.} Our current framework assumes a single delivery vehicle that executes deliveries in several subsequent subtours. This is because the MDP in the reinforcement learning pipelines of neural combinatorial optimization papers such as \cite{kool2018attention, kwon2020pomo, lischka2025greatarchitectureedgebasedgraph} builds solutions sequentially by adding one node at a time to a partial solution. Generalizing to several vehicles routed in parallel requires a different modeling approach, and whether the advantage of GREAT over PyVRP that we observe for larger instances carries over to such a setting remains an open empirical question. However, using only a single vehicle, the number of customers we can serve in a day is limited.

\paragraph{Interchangeable goods.} We assume that all goods are interchangeable, which allows the vehicle to serve a newly accepted customer with goods it already carries. This holds for pick-up operations such as parcel returns, but not for typical parcel delivery by large e-commerce platforms, where every customer receives specific goods. In such a setting, a dynamically arriving customer can only be served after the vehicle has reloaded at the depot. Within our framework, this can be incorporated by masking newly accepted customers in the decoder until the next depot visit, while the pricing logic remains unchanged. The model would, however, have to be trained with this additional masking.

\paragraph{Simulation and model specification.} While our Seattle-based road network results in realistic routing data, the customer data is mostly artificial. Customer locations reflect population density, however, demands, arrival times and choices are simulated. In particular, customer choices are generated by the same MNL model family that the bandit learns, with preferences shared by all customers. The learning problem is therefore well specified, which favors parameter recovery. With heterogeneous customers or a misspecified choice model, the learned parameters would only approximate the true behavior and learning might be slower. Further, dynamic pricing charges customers in sparsely served areas more for home delivery, which raises fairness questions we do not address. For a more realistic evaluation, real customer choice data is required. However, we believe the current evaluation can serve as a representative assessment of the potential of GREAT-ESDD.

 \section{Conclusion}
\label{sec:conclusion}

OOH delivery options can be an effective way to make delivery operations more efficient, provided that customers can be steered towards them. In this work, we studied how the operator of an online, same-day delivery service can use dynamic pricing to this end while customers order continuously and the delivery vehicle is already en route.

To achieve this, we build a framework that dynamically selects a price menu that aims to steer customer decisions and maximizes the expected profit. At the center of our work is a neural CVRP solver using GREAT as its encoder architecture, which we train to solve CVRP of different sizes and starting states. By this, we can call the solver any time a customer orders and solve the candidate CVRP instances of all home and OOH delivery options in parallel. We use the resulting increase in routing cost to select prices that trade off revenue and routing cost and, after the customer has made a decision, directly use the computed routing solution as the future plan of our delivery vehicle. Additionally, we learn unknown customer preferences online with Thompson sampling and a Laplace-approximated posterior.

Our experiments on the Seattle road network show that our neural solver is fast enough that customers are not kept waiting for a prolonged period of time, and that it outperforms a state-of-the-art metaheuristic (PyVRP) under a similar time budget, especially for larger instances. Within GREAT-ESDD, this results in an approximately 11\% higher daily profit than a PyVRP-based framework with the same decision time. Compared to a flat pricing strategy where every delivery option has the same price, GREAT-ESDD increases the OOH share from 18.6\% to 29.2\%, reduces the routing cost by approximately 6\% and lowers the average price paid by customers by 3.2\%, while increasing profit by approximately \$14 per day. When the customer preferences are unknown, the learned dynamic pricing outperforms flat pricing within 5 days of simulated operations. We conclude that fast neural routing makes it possible to price delivery options based on solved candidate routing problems at checkout, and that this cost-aware pricing pays off even when customer preferences have to be learned first.

GREAT-ESDD has limitations that we discuss in Section~\ref{sec:limitations}: the cost estimates are myopic and heuristic, the rectifier $\omega$ is static, and we consider a single vehicle with interchangeable goods and simulated customer behavior.

In future extensions, we plan to investigate how the value $\omega$ can be learned instead of being set to a static, predefined value. 
Potentially, an adaptive $\omega$ that changes with the (approximate) fraction of customers that have already ordered can improve the results further, since marginal routing cost estimations later in the day are more exact than earlier ones. 
Further directions include multiple vehicles, non-interchangeable goods, anticipatory (non-myopic) cost estimates, and validation with real customer choice data.

\section*{Acknowledgments}

This work was performed as a part of the research project ``LEAR: Robust LEArning methods for electric vehicle Route selection'' funded by the Swedish Electromobility Centre (SEC). The computations were enabled by resources provided by the National Academic Infrastructure for Supercomputing in Sweden (NAISS) at Chalmers e-Commons partially funded by the Swedish Research Council through grant agreement no. 2022-06725. We further acknowledge the University of Washington Valle Scholarship and Scandinavian Exchange Program for their support.

\printcredits

\bibliographystyle{cas-model2-names}

\bibliography{cas-refs}

\appendix

\section{$\omega$-value selection}
\label{app:omega_search}

\begin{figure}
    \centering
    \includegraphics[width=0.5\linewidth]{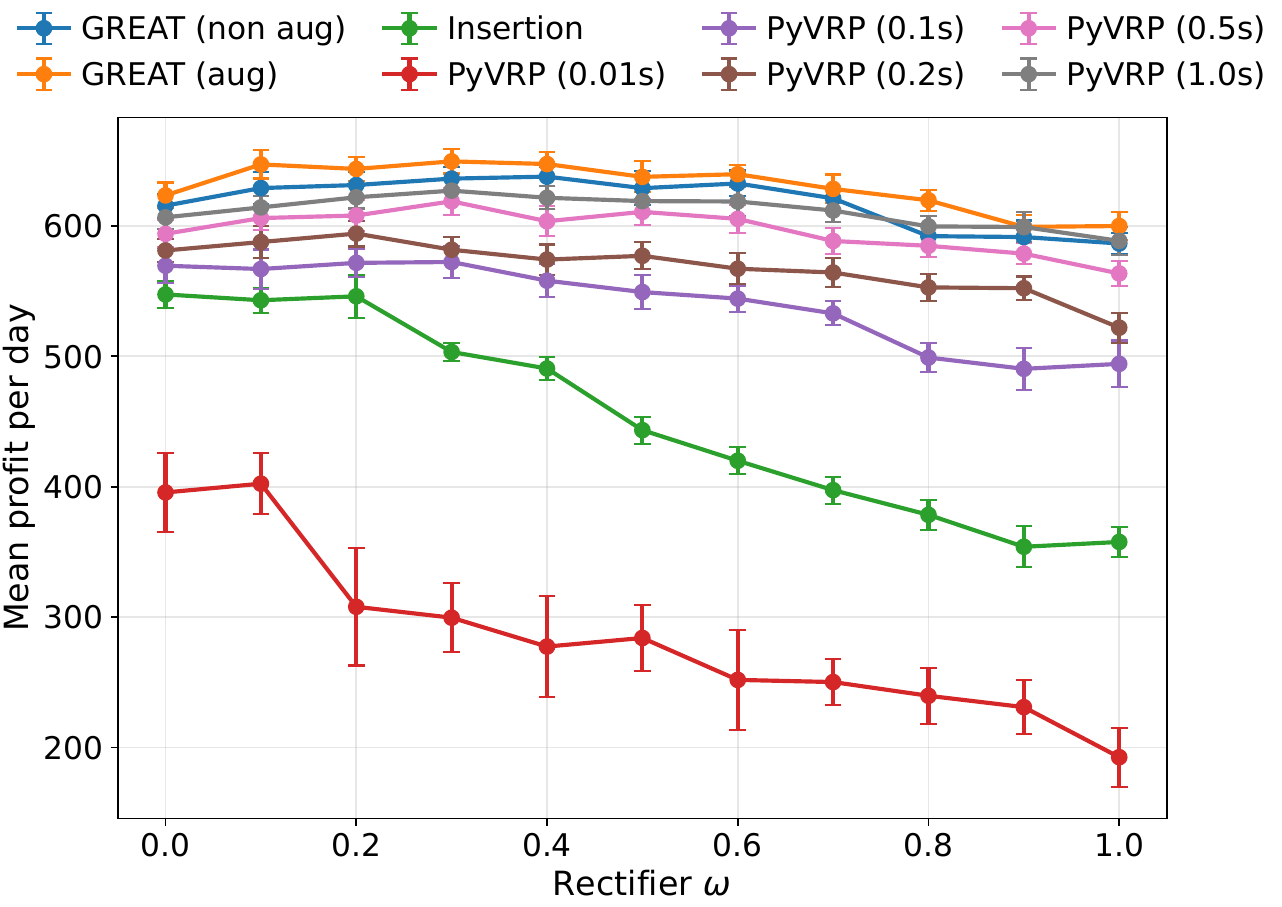}
    \caption{Average profit achieved by different solvers for different rectifiers $\omega$}
    \label{fig:scale_searhc}
\end{figure}

To estimate the best $\omega$ value of each routing method (``backend'') in the expected profit maximization, we simulate 10 days of operations using the different rectifiers in the set $\{0, 0.1, 0.2, \dots, 1\}$ and report the average profit.
The result can be observed in Figure~\ref{fig:scale_searhc}.
The best rectifiers for the different backends are as follows:
\begin{itemize}
    \item $\omega=0$ for the insertion heuristic
    \item $\omega=0.1$ for PyVRP with $0.01$s limit
    \item $\omega=0.2$ for PyVRP with $0.2$s limit
    \item $\omega=0.3$ for GREAT with augmentation, PyVRP with $0.1$s, $0.5$s and $1$s limit
    \item $\omega=0.4$ for GREAT without augmentation
\end{itemize}
However, we acknowledge that those values were computed by conducting a search on only 10 days, and is therefore susceptible to noise. Furthermore, since the grid of possible rectifiers was quite coarse, it is possible that better values can also be found by making it more fine-grained.
Nevertheless, for all experiments using scaling values $\omega$, we choose the best value according to this small ablation.

\section{Optimal price for flat pricing}
\label{app:flat_price}

\begin{figure}
    \centering
    \includegraphics[width=0.5\linewidth]{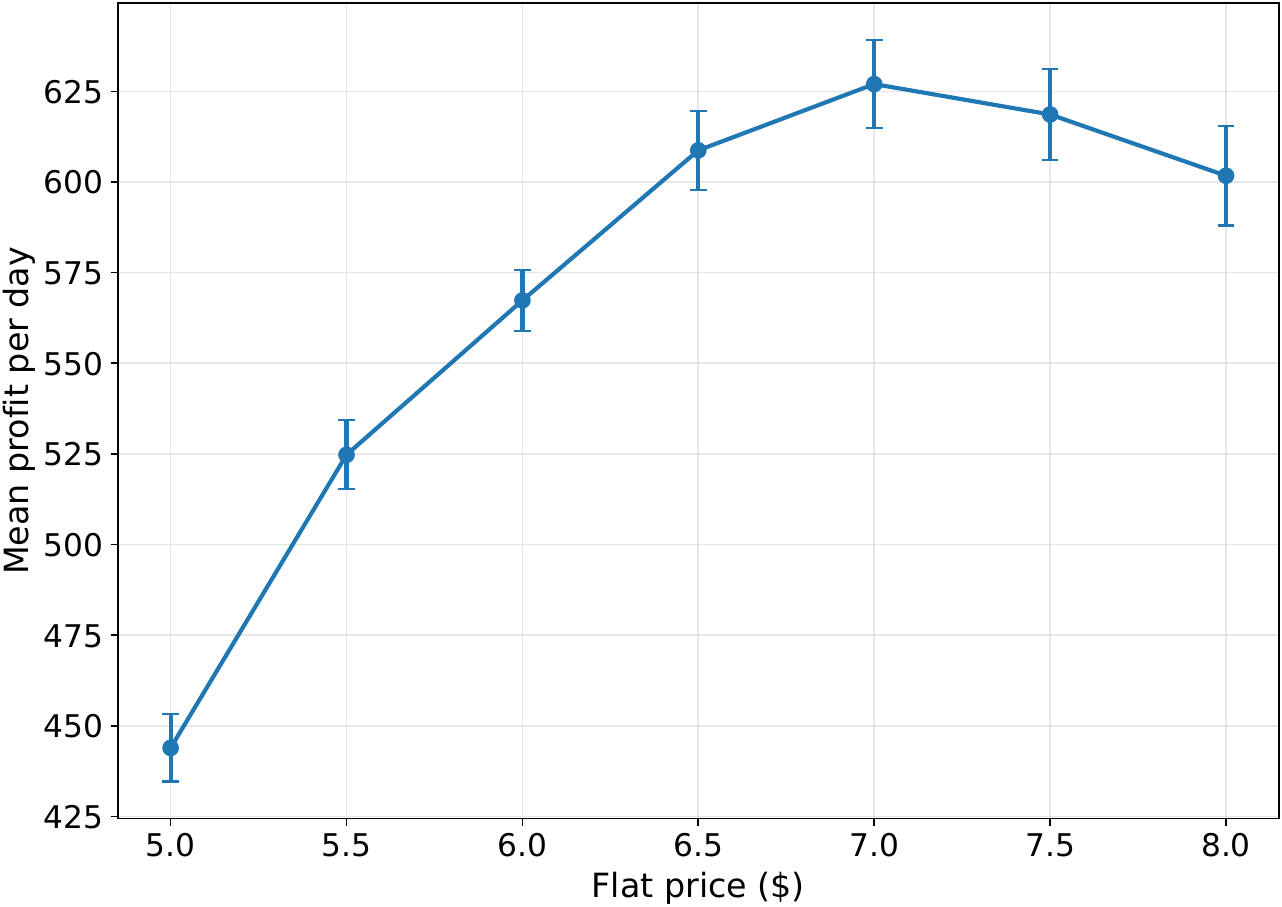}
    \caption{Profit development of flat pricing options}
    \label{fig:flat_price}
\end{figure}

As a simple pricing baseline, we can use flat pricing. That is, every customer gets the same price offered for all delivery options and there is no dynamic price adjustment. 
To this end, we run a small ablation where we simulate 10 days of operations using GREAT without augmentation as a backend to check which flat price results in the highest profit.
The result can be seen in Figure~\ref{fig:flat_price} where we show the profit of all prices in the range $\{5, 5.5, 6, \dots, 8\}$. 
The experiment shows that a flat price of $\$7$ leads to the highest profit for the company.

\section{Local search postprocessing}
\label{app:local_search}
We apply an optional, local-search-based postprocessing to the solutions generated by our trained GREAT model.
The procedure works as follows:
First, we segment a solution into its subtours between depot visits. 
Every such subtour is optimized independently, which leads to a very fast runtime, since a subtour usually contains approximately 10 nodes only, resulting in a greatly reduced search space (compared to optimizing a sequence over many more nodes). 
We note that this is a deliberate choice since we are not interested in applying a sophisticated, extensive search, but only want to improve local inefficiencies in the solution that possibly arise during the decoding procedure of the neural model. 
The aim of the search is not to re-optimize more complex, global patterns.
After the subtours have been split, we alternate between 2-opt \citep{croes1958method} and Or-opt \citep{or1976traveling} moves. 
We run this alternation up to 10 passes or until a local minimum is found (i.e., no further improvement is possible).
Both operators ensure the first and last node of the subtour stay the same to ensure the vehicle's current position (for a partially solved instance) and depot visits stay the same. 
Since we optimize the subtours individually and do not move customer visits between them, we can ensure that vehicle capacity constraints stay satisfied.

\end{document}